\documentclass[10pt,twocolumn,letterpaper]{article}

\usepackage{cvpr}      % To produce the REVIEW version
\usepackage[dvipsnames]{xcolor}

\definecolor{cvprblue}{rgb}{0.21,0.49,0.74}
\usepackage[pagebackref,breaklinks,colorlinks,citecolor=cvprblue]{hyperref}

\usepackage{amsmath,amssymb} % define this before the line numbering.
\usepackage{booktabs}
\usepackage{multirow}
\usepackage{algorithm}
\usepackage{algorithmic}

\usepackage{enumitem}
\usepackage{pifont}

\def\confName{CVPR}
\def\confYear{2024}

\title{WildFake: A Large-scale Challenging Dataset for AI-Generated Images Detection}
\author{\textnormal{Yan Hong}$^{1\ast}$, \textnormal{Jianming Feng}$^{1}$\thanks{Equal contribution.},  \textnormal{Haoxing Chen}$^{1}$ \\ \textnormal{Jun Lan}$^{1}$, \textnormal{Huijia Zhu}$^{1}$, \textnormal{Weiqiang Wang}$^{1}$, \textnormal{Jianfu Zhang}$^{2}$\thanks{Corresponding author.}~ \\
$^{1}$ Ant Group, 
$^{2}$ Qing Yuan Research Institute, Shanghai Jiao Tong University \\
$^{1}$ {\tt\small yanhong.sjtu@gmail.com, hx.chen@hotmail.com} \\
$^{1}$ {\tt\small \{fengji.fjm, yelan.lj, huijia.zhj, weiqiang.wwq\}@antgroup.com} \\
$^{2}$ {\tt\small c.sis@sjtu.edu.cn}
}

\begin{document}
\maketitle
\begin{abstract}
The extraordinary ability of generative models enabled the generation of images with such high quality that human beings cannot distinguish Artificial Intelligence (AI) generated images from real-life photographs.  
The development of generation techniques opened up new opportunities but concurrently introduced potential risks to privacy, authenticity, and security.
Therefore, the task of detecting AI-generated imagery is of paramount importance to prevent illegal activities.
To assess the generalizability and robustness of AI-generated image detection, we present a large-scale dataset, referred to as WildFake, comprising state-of-the-art generators, diverse object categories, and real-world applications.
WildFake dataset has the following advantages: 
1) Rich Content with Wild collection: WildFake collects fake images from the open-source community, enriching its diversity with a broad range of image classes and image styles. 
2) Hierarchical structure: WildFake contains fake images synthesized by different types of generators from GANs, diffusion models, to other generative models. 
These key strengths enhance the generalization and robustness of detectors trained on WildFake, thereby demonstrating WildFake's considerable relevance and effectiveness for AI-generated detectors in real-world scenarios. 
Moreover, our extensive evaluation experiments are tailored to yield profound insights into the capabilities of different levels of generative models, a distinctive advantage afforded by WildFake's unique hierarchical structure.
\end{abstract}

\begin{table*}
\centering
\caption{
Comparison among our WildFake dataset and existing fake image detection datasets. 
% In ``Community'' column represents fake images generated by our generation pipeline (\emph{resp.,} fake images collected from the open-source community). The 'Hier' denotes 'hierarchical' which means the datasets whether is classified into fine-grained subsets for cross-validation.
% For content column, 'Closed' (\emph{resp.,} 'Open') means that fake images are produced based on the fixed datasets (\emph{resp.,} fake images collected or produced based a open-set label or caption set).
} \label{tab:dataset_compare}
\resizebox{2.0\columnwidth}{!} {
\begin{tabular}{l|c|c|c|c|c|c|c|c}
\toprule
\multirow{2}{*}{\textbf{Datasets}}&
\multicolumn{3}{c|}{\textbf{Generators}}&
\multicolumn{1}{c|}{\multirow{2}{*}{\textbf{Communities}}} &
\multicolumn{1}{c|}{\multirow{2}{*}{\textbf{Available}}}&
\multicolumn{1}{c|}{\multirow{2}{*}{\textbf{Hierarchies}}} &
\multicolumn{2}{c}{\textbf{Image Numbers}} \\
\cline{2-4} \cline{8-9}  
&\textbf{GANs} & \textbf{DMs} &\textbf{Others}  &\multicolumn{1}{c|}{}  & \multicolumn{1}{c|}{} & \multicolumn{1}{c|}{} & \textbf{Fake} &\textbf{Real} \\
\midrule
CNNSpot \cite{wang2020cnn} & \ding{51} & \ding{55} & \ding{55}     &  \ding{55} & \ding{51} & \ding{55} & 362,000 & 362,000\\
IEEE VIP Cup \cite{verdoliva2022} & \ding{51} & \ding{51} & \ding{55}   &  \ding{55} & \ding{55} & \ding{55} & 7,000 & 7,000\\
DE-FAKE \cite{sha2022fake} & \ding{55} & \ding{51} & \ding{55}   &  \ding{55} & \ding{55} & \ding{55} & 20,000 & 60,000\\
CiFAKE \cite{bird2023cifake} & \ding{55} & \ding{51} & \ding{55}  &  \ding{55} & \ding{51} & \ding{55} & 60,000 & 60,000\\
GenImage~\cite{zhu2023genimage} & \ding{51} & \ding{51} & \ding{55}  &  \ding{55} & \ding{51} & \ding{55} & 1,331,167 & 1,350,000 \\
DiffusionDB~\cite{wang2022diffusiondb} & \ding{55} & \ding{51} & \ding{55}  &  \ding{51} & \ding{51} & \ding{55} & 14,000,000 & 0 \\
ArtiFact~\cite{rahman2023artifact} & \ding{51} & \ding{51} & \ding{55} &\ding{55} & \ding{51} & \ding{55}  & 1,521,900 & 962,200 \\
DiffusionForensics~\cite{wang2023dire} &\ding{55} & \ding{51} & \ding{55}& \ding{55}&  \ding{51} & \ding{55} & 439,020&92,000   \\
WildFake & \ding{51} & \ding{51} & \ding{51} &\ding{51}  & \ding{51} & \ding{51} & 2,557,278 & 1,013,446  \\
\bottomrule
\end{tabular}
}
\end{table*}

\input{Sections/Tables/data_distribution}

\section{Introduction}
The development of generative models has markedly improved the creation of realistic images, simplifying the process of producing AI-generated images (\textit{i.e.}, fake images). 
This heightened accessibility has amplified concerns regarding the widespread spread of false information.
AI-generated images, characterized by their striking visual clarity, are particularly persuasive and have the potential to significantly influence public opinion in critical domains such as politics and economics.
% Altered images are especially persuasive due to their visual clarity, which can influence public opinion in areas like politics and economics.
% Generative models can be coarsely classified into three categories including diffusion-based models, gan-based models, and other types.
% Recent diffusion models~\cite{saharia2022photorealistic,rombach2022high,} have gained immense popularity by enabling high-quality and controllable image generation based on text prompts written in natural language~\cite{rombach2022high}. Recent generative adversarial networks~\cite{alanov2023styledomain,sauer2022stylegan,pehlivan2023styleres,karras2019style,esser2021taming,choi2020stargan,brock2018large,tao2023galip,tao2022df,kang2023scaling} also make significant progress in improving the quality of synthesized images. Besides, 
% Since the release of these models, people from different domains have quickly applied them to create award-winning artworks~\cite{everaert2023diffusion,hamazaspyan2023diffusion,zhang2023inversion}, synthetic radiology images~\cite{zhang2020radiology}, and even hyper-realistic videos~\cite{ho2022imagen,yang2023diffusion,singer2022make}. 
To counteract such harmful activities, the development of technologies capable of detecting altered images is essential. 
Generative models typically imprint unique patterns, which don't appear in real images and vary depending on the model and its training data~\cite{marra2019gans}.
Recent research in synthetic image detection has focused on identifying these irregularities through methods like color pattern analysis, light intensity evaluation, and Fourier spectrum analysis~\cite{corvi2023detection,frank2020leveraging}.
While traditional techniques based on manually selected features and frequency analysis show limited efficacy, deep CNN models are more effective in pattern detection~\cite{marra2018detection}. 
However, with the range of models available—from Generative Adversarial Networks (GANs) \cite{alanov2023styledomain,sauer2022stylegan,pehlivan2023styleres,karras2019style,esser2021taming,choi2020stargan,brock2018large,tao2023galip,tao2022df,kang2023scaling} to Diffusion Models (DMs) \cite{saharia2022photorealistic,rombach2022high}, users can now easily produce high-quality and diverse images with different types of model with personalized weights, and spread those generated images with different social media platforms. 
These images, often disseminated across various social media platforms, present a significant challenge in terms of generalization and robustness for detection technologies. Existing detectors still face difficulties with generative models not encountered during their training phase \cite{aghasanli2023interpretable,wu2023generalizable,lorenz2023detecting,wang2023dire}.

\begin{figure*}[t]
\begin{center}
\includegraphics[width=1.0\linewidth]{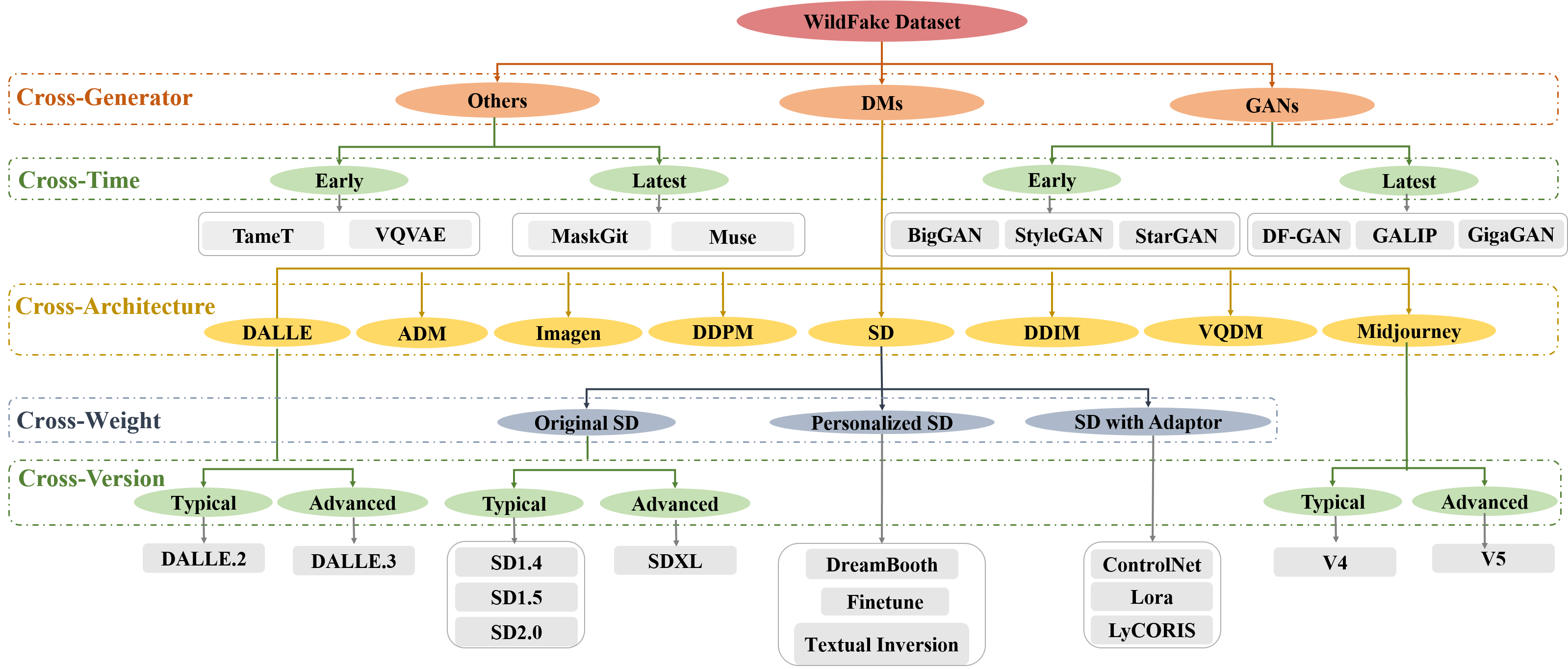}
\end{center}
 \caption{Overview of WildFake dataset. (a) At the cross-generator level, we separate generators into DMs, GANs, and Others. (b) The cross-architecture level discriminates different architectures from DMs, \emph{e.g.,} DALLE~\cite{ramesh2022hierarchical}, ADM~\cite{dhariwal2021diffusion}, Imagen~\cite{saharia2022photorealistic}, DDPM~\cite{ho2020denoising}, DDIM~\cite{song2020denoising}, VQDM~\cite{gu2022vector}, Midjouney~\cite{Midjourney}, and SD~\cite{rombach2022high}. Then, we separate fake images from SD~\cite{rombach2022high} into three subsets according to the cross-weight level. We also introduce a cross-version level to separate different generators into typical classes and advanced classes. More the dataset details can be found in the supplementary material.}
% \vspace{-8pt}
\label{fig:framework_dataset} 
\end{figure*}

To aid in the development of detectors, many datasets for general AI-generated images are built 
\cite{wang2020cnn,verdoliva2022,sha2022fake,bird2023cifake,wang2022diffusiondb,rahman2023artifact,zhu2023genimage,wang2023dire}.
However, these existing datasets often exhibit significant limitations. They are generally restricted to one or two types of generators, constrained to producing images within fixed categories, or largely dependent on low-quality, user-generated images. These constraints hinder the effectiveness and adaptability of detectors in recognizing a broader range of AI-generated content.
In this paper, we present WildFake, a comprehensive, large-scale dataset specifically designed for the detection of AI-generated images.
We summarize the comparison among fake image detection datasets in ~\cref{tab:dataset_compare}.
WildFake stands out by generating a diverse array of rich, stylistically varied, and high-quality images. 
Within the WildFake dataset, fake images are produced either through our extensive generation pipeline or sourced from open-source communities, where users share images created with their personalized generative models. To augment the dataset's diversity, real images are gathered from open datasets used in various tasks like image captioning, generation, and classification, ensuring a broad spectrum of styles and content.
We have conducted a series of experiments on the WildFake dataset to assess the generalization capabilities of detectors trained on fake images, demonstrating WildFake's potential in enhancing the understanding of fake image detection in a multitude of real-world scenarios. Additionally, we have implemented a series of degradation tests on the WildFake testing set, illustrating the robustness of these detectors in challenging conditions.  
Besides, Distinct from existing datasets, WildFake categorizes generative models into three primary groups: GANs\cite{karras2019style,karras2020analyzing,karras2021alias,choi2018stargan,choi2020stargan,brock2018large,kang2023scaling,tao2023galip}, DMs \cite{gu2022vector,Midjourney,ho2020denoising,song2020denoising,nichol2022glide,ramesh2022hierarchical,rombach2022high,dalle3}, and Others \cite{van2017neural,esser2021taming,chang2022maskgit,chang2023muse}. 
Based on these methods, WildFake uniquely features four levels of categorization, each based on different dimensions, as depicted in \cref{fig:framework_dataset}.
WildFake comprehensively incorporates multiple generator types, various architectures, different model weights, and versions of the same model series. Such a structure is conducive to a detailed analysis of various image generators, offering insights of their characteristics.
We plan to publicly release the WildFake dataset to support and aid the academic community.

% \input{Sections/Figures/real_data_distribution}
% \input{Sections/Figures/fake_data_distribution}

% In \cref{fig:framework_dataset}

\section{Related Works}
In this section, we offer a brief yet thorough exploration the image generation methods ( \cref{sec:image_gen}), existing AI-generated image datasets (\cref{sec:other_datasets}), and existing AI-generated image detection approaches \cref{sec:image_detectors}. 

\subsection{Image Generation} \label{sec:image_gen}
Raging from Generative Adversarial Networks (GANs)~\cite{alanov2023styledomain,sauer2022stylegan,pehlivan2023styleres,karras2019style,esser2021taming,choi2020stargan,brock2018large,tao2023galip,tao2022df,kang2023scaling} to DMs (DMs) ~\cite{saharia2022photorealistic,rombach2022high}, a multitude of users can now effortlessly produce images of high quality and diversity using varied models with personalized weights, and spread those generated images across various social media platforms. 
Besides, other researchers are dedicated to leveraging the autoregression models with vector quantization techniques~\cite{esser2021taming,van2017neural,huang2023towards,razavi2019generating} or masked transformers~\cite{li2023mage,razavi2019generating,huang2023not} for high-quality image synthesis. 
We propose a tripartite classification of these generative models into: \textit{GANs}, \textit{DMs}, and \textit{Others}. 

\noindent{\textbf{GANs for Image Generation:}}
GAN~\cite{goodfellow2014generative} was proposed to discriminate real samples from fake samples and generate more realistic samples. In the early stage, unconditional GANs: Wasserstein-GAN~\cite{arjovsky2017wasserstein}, StyleGANs~\cite{karras2019style,karras2020analyzing,karras2021alias}, and BigGAN~\cite{brock2018large} use random vectors to generate realistic samples based on the learned distribution of training samples. 
Then, GANs conditioned on input single image are proposed~\cite{antoniou2017data,ZhangHLZ019}.
This transition led to the development of models like CycleGAN~\cite{zhu2017unpaired} and StarGAN~\cite{choi2018stargan,choi2020stargan}, which transform a given image into a desired target image through adversarial training.
Recently, text-conditioned GANs, such as DF-GAN~\cite{tao2022df} and GALIP~\cite{tao2023galip} were trained on pairwise text-image data and generate new images from textual descriptions. 
Moreover, GigaGAN~\cite{kang2023scaling} has introduced a novel GAN framework, showcasing the potential of GANs in the text-to-image synthesis domain.

\noindent{\textbf{DMs for Image Generation:}}
DMs generators~\cite{ho2020denoising,song2020denoising} have recently achieved remarkable performances in image synthesis.
DDPM~\cite{ho2020denoising}  establishes a sequence of diffusion steps in a Markov chain, progressively infusing Gaussian noise into the data until it converges to an isotropic Gaussian distribution in a forward process, and subsequently learns to inversely reconstruct the original data from the noise in a reverse process.
DDIM~\cite{song2020denoising} proposes a new deterministic method for accelerating the iterative process without the Markov hypothesis. 
ADM~\cite{dhariwal2021diffusion} finds a much more effective architecture and further achieves outstanding performance with the integration of classifier guidance.
Stable Diffusion~\cite{rombach2022high} (SD) is an advanced text-to-image diffusion model capable of generating highly realistic images based on any given text input. 
Different versions of SD including SD1.4, SD1.5, SD2.0, and SD-XL trained on large-scale datasets like Laion5B and Laion-Aesthetics dataset~\cite{Laionaesthetics} with different dataset selection strategies to improve the quality of generated images and to ensure the safe generation of content. 
DALLE.2~\cite{ramesh2022hierarchical} align visual embeddings to textual embeddings for supporting text-to-image generation and image-to-image generation, and DALLE.3~\cite{dalle3} adopt advanced caption system to improve the quality of training image-text pairs. 
VQDM~\cite{gu2022vector} proposes a latent-space method that eliminates the unidirectional bias with previous methods and incorporates a mask-and-replace diffusion mechanism to alleviate the accumulation of errors. 
Imagen~\cite{saharia2022photorealistic} adopt a generic large language model pre-trained on text-only corpora to generate high-fidelity images. 
Beyond these academic contributions, Midjourney~\cite{Midjourney} is one of the most renowned commercial software programs, known for its exceptional image generation performance. 
Our implementation employs Midjourney V5 for image generation, which offers more intricate details compared to previous versions, yielding images that closely resemble real-world photographs. 

\begin{figure*}[t]
\begin{center}
\includegraphics[width=1.0\linewidth]{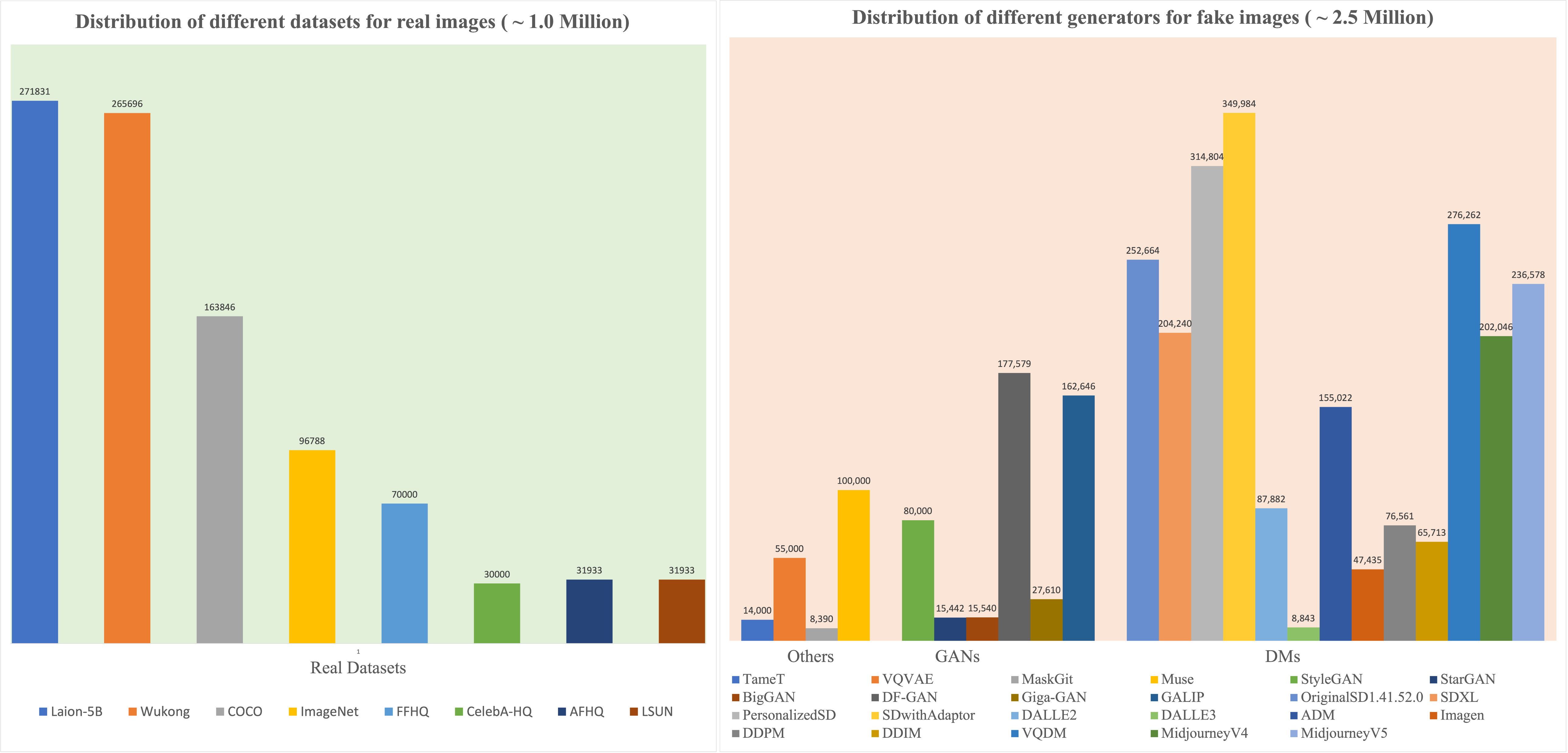}
\end{center}
 \caption{Overview of data distribution of WildFake dataset. The left subfigure shows the distribution of real images from open-source datasets ,and the right subfigure represents fake images sourced from different generators.}
% \vspace{-8pt}
\label{fig:data_distribution} 
\end{figure*}

\noindent{\textbf{Other Image Generation Approaches:}}
In addition to DMs and GANs, other generative models also demonstrate potential for high-quality image synthesis and offer valuable insights for theoretical analysis. These include Variational Auto-Encoders (VAEs) \cite{kingma2013auto,doersch2016tutorial,higgins2016beta} and Flow-Based Models \cite{rezende2015variational,dinh2016density,kingma2018glow}.
Recent developments have further demonstrated the capability of other type of generative models in producing high-quality images. 
VQVAE~\cite{van2017neural} introduce vector quantization technique into VAEs~\cite{bao2017cvae} to solve vague problems in VAEs.  
Vector quantization~\cite{van2017neural} is also combined with autoregression model in TameT~\cite{esser2021taming} to produce high-resolution images. 
Masked Generative Encoder~\cite{li2023mage,he2022masked} represents a convergence of masked generative models and self-supervised representation learning. This union forms a robust framework for image synthesis, leveraging the strengths of both paradigms to enhance the generative process.
MaskGit \cite{chang2022maskgit} introduces a bidirectional transformer decoder. This innovative mechanism enables simultaneous token generation for an entire image, which is then progressively refined in an iterative manner, predicated on the tokens generated in the preceding iteration.
Muse \cite{chang2023muse}, a text-to-image Transformer model, is also noteworthy. It is trained on a masked modeling task in discrete token space, utilizing text embeddings extracted from a pre-trained large language model to achieve impressive image generation performance.
% MAE~\cite{he2022masked} developed an asymmetric encoder-decoder architecture, with an encoder that operates only on the visible subset of patches, along with a lightweight decoder that reconstructs the original image from the latent representation and mask tokens.

% \begin{figure}[t]
% \begin{center}
% \includegraphics[width=1.0\linewidth]{./Figures/other_based.png}
% \end{center}
%  \caption{Visualization of some examples from Other-based generators in the WildFake dataset.}
% % \vspace{-8pt}
% \label{fig:vis_other} 
% \end{figure}

% \begin{figure}[t]
% \begin{center}
% \includegraphics[width=1.0\linewidth]{./Figures/GAN_based.png}
% \end{center}
%  \caption{Visualization of some examples from GAN-based generators in the WildFake dataset.}
% % \vspace{-8pt}
% \label{fig:vis_gan} 
% \end{figure}

% \begin{figure*}[t]
% \begin{center}
% \includegraphics[width=1.0\linewidth]{./Figures/Diffusion_based.png}
% \end{center}
%  \caption{Visualization of some examples from Diffuion-based generators in the WildFake dataset.}
% % \vspace{-8pt}
% \label{fig:vis_diffusion} 
% \end{figure*}

% \begin{figure*}[t]
% \begin{center}
% \includegraphics[width=1.0\linewidth]{./Figures/real_based.png}
% \end{center}
%  \caption{Visualization of some examples from real datasets in the WildFake dataset.}
% % \vspace{-8pt}
% \label{fig:vis_real} 
% \end{figure*}

\begin{figure*}[t]
\begin{center}
\includegraphics[width=0.9\linewidth]{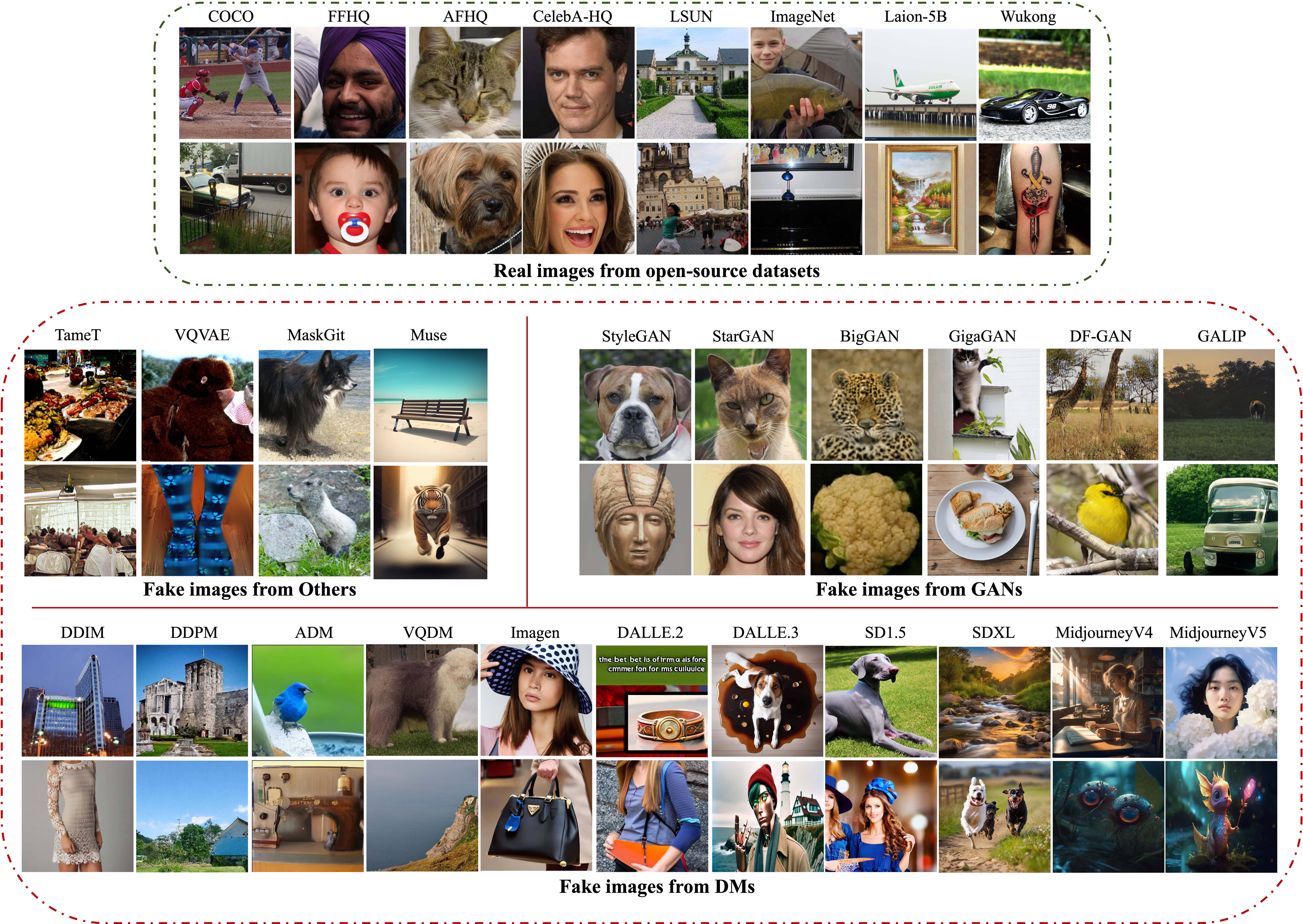}
\end{center}
 \caption{Visualization of some real images and fake images from the WildFake dataset.}
% \vspace{-8pt}
\label{fig:visualization} 
\end{figure*}

\subsection{AI-Generated Image Datasets} \label{sec:other_datasets}
To enhance the development of AI-generated image detectors, early datasets primarily utilized GANs. 
For instance, CNNSpot~\cite{wang2020cnn} builds its foundation on GAN-generated images. This dataset exclusively uses ProGAN~\cite{karras2017progressive} to generate its training set and assesses the performance of detectors across various GAN-based testing sets.
However, the recent advent of alternative generators besides GANs, such as DMs, has significantly improved the quality of generated images. 
his advancement complicates the differentiation between real and synthetic images. Consequently, investigating images produced by DMs becomes crucial.
In response to this need, IEEE VIP Cup~\cite{verdoliva2022} and DE-FAKE~\cite{sha2022fake} have incorporated DMs to create more diverse images, using prompts from the MSCOCO dataset~\cite{lin2014microsoft} and Flickr30k dataset~\cite{young2014image}. 
Based on a small-scale Cifar10 dataset~\cite{krizhevsky2009learning}, CiFAKE~\cite{bird2023cifake} generates fake images only with Stable Diffusion V1.4~\cite{rombach2022high}. 
DiffusionDB~\cite{wang2022diffusiondb}, constructed by collecting images shared on the Stable Diffusion public Discord server~\cite{website2023}, does not categorize DiffusionDB dataset based on different dimensions like model architectures and parameters. This lack of differentiation hinders the evaluation of detectors using multi-dimensional metrics.
ArtiFact~\cite{rahman2023artifact} generates fake images using real prompts from existing datasets. However, this approach limits the diversity and creativity of the fake images, as they are constrained by the originality of the prompts.
GenImage~\cite{zhu2023genimage} also faces limitations, relying on the $1000$ classes from ImageNet dataset~\cite{russakovsky2015imagenet} to produce limited content. It is not suited to real-world scenarios where users create fake images using a variety of random textual information.
DiffusionForensics dataset is proposed in DIRE~\cite{wang2023dire} which adopt several DMs to produce fake images based on LSUN-Bedroom~\cite{yu2015lsun}, ImageNet~\cite{russakovsky2015imagenet}, and CelebA-HQ~\cite{karras2017progressive} dataset. 

Current datasets in this field tend to focus on a limited range of image generators, resulting in a narrow representation of fake images. These datasets often feature synthetic images either created by the developers themselves or sourced from a handful of open-source platforms, thus offering a restricted view of the diversity and intricacy inherent in fake imagery. 
In contrast, our study adopts a broader approach, systematically collecting and generating fake images from various open-source websites and diverse generators according to a hierarchical construction structure. This method offers a more comprehensive perspective on the complexity of AI-Generated Image Detection.

\subsection{AI-Generated Image Detectors} \label{sec:image_detectors}
A naive method is to establish a classification model to distinguish fake images from real images via finetuning pre-trained discriminative models such as ViT~\cite{parmar2018image} or ResNet~\cite{he2016deep} for fake vs. real image classification.
DIRE~\cite{wang2023dire} is mainly designed to distinguish images generated by DMs models from real images by measuring the error between an input image and its reconstruction counterpart by a pre-trained diffusion model.
IFDL~\cite{guo2023hierarchical} introduces a multi-branch system combining feature extraction, localization, and classification to assign multi-level labels to fake images. 
multiLID~\cite{lorenz2023detecting} eemploys feature-map representations of fake images and calculates multi-local intrinsic dimensionality for classifier training. 
LASTED~\cite{wu2023generalizable} formulates synthetic image detection as an identification problem by augmenting the training images with carefully designed textual labels with joint image-text contrastive learning.

\section{Dataset Construction}
Addressing the critical need for assessing the generalizability and robustness of fake image detectors, we have developed the ``WildFake'' dataset, which meets two principal criteria:
\begin{itemize}[leftmargin=*]
    \item Diverse Content with Wild Collection:  WildFake includes a wide array of high-quality fake images sourced from open-source websites, along with images produced using both user-trained and officially provided pre-trained generative models. This diverse collection ensures a comprehensive set of fake images, significantly enriching the understanding of fake image detection across numerous real-world contexts.
    \item Hierarchical Structure: The dataset is organized hierarchically, encompassing cross-generators, cross-architectures within the same type of generator, cross-weights within identical architectures, and cross-time analysis either within the same generator type or across different versions of the same model series. This structure facilitates in-depth analysis of various image generators.
\end{itemize}

% The challenge of evaluating the performance of fake image detectors in terms of their generalizability and robustness requires a comprehensive dataset that fulfills several criteria: 
% obtaining high-quality fake images collected from open-source websites and producing new fake images with user-trained generative models or officially provided pre-trained generative models. 
% fake images from a variety of coarse categories or fine-grained texts
% Taking the above-mentioned requirements into consideration, we propose a large-scale challenging dataset, named WildFake, which adopts a hierarchical structure to collect fake images from a variety of generators, diverse content, and real-world scenes, providing researchers with a more comprehensive understanding of fake image detectors. 

\subsection{Organization of WildFake Dataset}
As is shown in ~\cref{fig:framework_dataset}, the proposed WildFake dataset consists of five levels consisting of \textit{cross-generator}, \textit{cross-architecture}, \textit{cross-weight}, \textit{cross-time}, and \textit{cross-version}.
\begin{itemize}[leftmargin=*]
    \item Cross-Generator: 
    This level encompasses DMs, GANs, and Other generators, providing a comprehensive overview of the diverse generative models in use.
    \item Cross-Time: 
    Focusing on GANs and Other generators known for high-quality synthesis, we categorize them into ``Early'' and ``Latest'' groups. ``Early'' represents well-established, popular models, whereas ``Latest'' includes recent advancements.
    Early GANs (\emph{advanced GANs}) consists of BigGAN~\cite{brock2018large}, StyleGANs~\cite{karras2019style,karras2020analyzing,karras2021alias}, and StarGANs~\cite{choi2018stargan,choi2020stargan}, (\emph{resp.,} GigaGAN~\cite{kang2023scaling}, DF-GAN~\cite{tao2022df}, and GALIP~\cite{tao2023galip}). Similarly, for Others generators, ``Early'' includes VQVAE~\cite{van2017neural} and VQGAN~\cite{esser2021taming}, while ``Latest'' encompasses Muse~\cite{chang2023muse} and Maskgit~\cite{chang2022maskgit}.
    \item Cross-Architecture:
    Considering the rapid development of DMs generators, nine kinds of DMs generators comprise cross-architecture level, consisting DALLE~\cite{ramesh2022hierarchical}, ADM~\cite{dhariwal2021diffusion}, Imagen~\cite{saharia2022photorealistic}, DDPM~\cite{ho2020denoising}, DDIM~\cite{song2020denoising}, VQDM~\cite{gu2022vector}, Midjouney~\cite{Midjourney}, and SD~\cite{rombach2022high}.  
    \item Cross-Weight:
    Open-source SD~\cite{rombach2022high} has been widely spread in academia and industry, officially released pre-trained models armed with updated architecture trained on large datasets, and users also adopt different finetuning strategies such as finetuning several modules of SD~\cite{rombach2022high} or finetuning with DreamBooth~\cite{ruiz2023dreambooth} to obtain personalized models. Besides, many works focus on training different adaptors to combine with the base SD model to achieve controllable generation. ControlNet~\cite{zhang2023adding} relies on paired image-prior data to control different prior information of generated images like edge, segmentation mask, style, and etc. Lora-based methods including Lora~\cite{hu2021lora} and LyCORIS~\cite{yeh2023navigating} are also proposed to train extra low-rank layers to incorporate new content into the base model. Also, there are some methods~\cite{gal2022image, voynov2023p+} to learn new tokens on the user-provided data for customized image generation. Thus, we classify SD-based generators into Original SD, Personalized SD, and SD with adaptors for cross-weight evaluation. % shown in ~\cref{fig:framework_dataset}.
    \item Cross-Version: 
    DALLE~\cite{ramesh2022hierarchical}, Midjouney~\cite{Midjourney}, Imagen~\cite{saharia2022photorealistic}, and SD~\cite{rombach2022high} have been widely known in academia and industry, due to the superiority of the quality of generated images. Fake images in DALLE~\cite{ramesh2022hierarchical} (\emph{resp.,} Midjouney~\cite{Midjourney}) are divided into ``Typical'' and ``Advanced'' subsets along the cross-version level. 
\end{itemize}

% DMs generators consist of DDIM~\cite{song2020denoising},
% DDPM~\cite{ho2020denoising},
% VQDM~\cite{gu2022vector}, ADM~\cite{dhariwal2021diffusion}, Imagen~\cite{saharia2022photorealistic}, DALLE~\cite{ramesh2022hierarchical},
% SD~\cite{rombach2022high}
% VQDM~\cite{gu2022vector}, and Midjourney~\cite{Midjourney}.
% MAE~\cite{he2022masked},

% \subsection{Fake Image Generation}
\subsection{Image Collection}
\noindent\textbf{Fake Image Collection:} 
To collect diverse fake images from different resources, we have established a generation pipeline. This pipeline facilitates the production of images using popular generative mechanisms, including GANs, DMs, and Others generators. 
Besides, we sourced user-created images from open-source platforms such as Civitai~\cite{civitai} and Midjourney~\cite{Midjourney}. 
On these platforms, users generate new images using either original open-source models or personalized models fine-tuned with their data. 
Unlike datasets primarily composed of author-generated images, such as  DiffusionForensics~\cite{wang2023dire}, ArtiFact~\cite{rahman2023artifact}, and GenImage~\cite{zhu2023genimage}, our approach of collecting from open sources offers a more representative sample of the average quality of generated images. This ensures that the evaluation of detection models on our dataset is reflective of real-world applicability.
For gathering images from GANs and Others, we primarily utilize official GitHub repositories and model cards from Hugging Face. When these GitHub repositories include generated samples, we directly extract fake images from there. In cases where the methods are associated with text-to-image  generation, new images are produced by randomly sampling captions from their respective testing datasets. For other scenarios, we randomly generate images using the pretrained models available.

\noindent\textbf{Real Image Collection:} 
Considering the fact that some fake images from GANs and Others are limited to specific domains determined by training datasets such as COCO~\cite{lin2014microsoft}, FFHQ~\cite{karras2019style}, ImageNet~\cite{deng2009imagenet}, LSUN Church~\cite{yu2015lsun}, CelebA-HQ~\cite{karras2017progressive}, AFHQ~\cite{choi2020stargan} dataset, we sample parts of real images from those datasets. 
Besides, recent text-to-image generators mostly trained on Laion-5B~\cite{schuhmann2022laion} or Chinese cross-modal 
Wukong~\cite{gu2022wukong} datasets, we also include real image samples from these text-to-image datasets. 
This ensures a comprehensive collection of real images, facilitating a more robust and realistic evaluation of image authenticity.

\begin{table*}[h]
\centering
\caption{Evaluating the superiority of our proposed WildFake dataset. ResNet50-ArtiFact (\emph{resp.,} ResNet50-WildFake) denotes the detector with ResNet50 architecture trained on ArtiFact (\emph{resp.,} WildFake dataset).  ViT-ArtiFact (\emph{resp.,} ViT-WildFake) denotes the detector with ViT architecture trained on ArtiFact, dataset (\emph{resp.,} WildFake dataset). (ACC($\%$), AP($\%$), and AUC($\%$)) are reported. 
%(ACC/AP/AUC in the Table)
} \label{tab:evaluation_superority}
\resizebox{2.05\columnwidth}{!} {
\begin{tabular}{c|c|c|c|c|c|c}
\toprule
\multirow{2}{*}{Training Detectors and Datasets}&
\multicolumn{5}{c|}{Testing Dataset} &
\multicolumn{1}{c}{\multirow{2}{*}{Avg}} \\
\cline{2-6}
  & DiffusionForensics  & GenImage & DiffusionDB & ArtiFact & WildFake & \multicolumn{1}{c}{} \\
\midrule
ResNet50-ArtiFact & 85.4/94.9/76.7  &76.5/84.8/82.9 & 64.1/69.9/68.1 &\textbf{97.2/99.5/99.3}& 85.4/94.9/76.7 & 81.7/88.8/80.74  \\

ResNet50-WildFake &\textbf{87.2/96.6/83.4}  & \textbf{80.9/89.9/89.3} &\textbf{96.3/99.2/ 99.2} &68.0/84.7/75.3 & \textbf{99.6/99.9/99.9} & \textbf{86.4/94.1/89.42} \\
\midrule

ViT-ArtiFact & 84.2/96.1/82.5  &78.5/88.1/85.0 & 68.4/75.3/73.2 &\textbf{96.8/99.6/99.5} & 84.2/96.1/82.5  & 82.4/91.0/84.4\\

ViT-WildFake & \textbf{95.8/99.1/97.2}  & \textbf{88.6/83.6/89.7} &\textbf{99.3/99.8/99.9} & 62.2/81.9/68.8 &\textbf{99.1/99.9/99.9} &\textbf{89.0/92.84/91.1} \\
\bottomrule
\end{tabular}
}
\end{table*}

\subsection{Analyses of WildFake}
Examples of images from WildFake's various categories are displayed in  ~\cref{fig:visualization}.
To analyze WildFake, we illustrate the distribution of both real and fake images from various sources in ~\cref{fig:data_distribution} for fake images and ~\cref{fig:data_distribution} for real images.
The WildFake dataset contains a total of 3,694,313 images, comprising 1,013,446 real images and 2,680,867 fake images. 
We split real images (\emph{resp.,} fake images) into the training set and testing set as the ratio of $4:1$. 
In detail, for all generators in ~\cref{fig:data_distribution},  $20\%$ samples are randomly selected as the testing set from fake images generated by each generator, with the remainder forming the training set. 
A similar splitting strategy is applied to the real datasets shown in  ~\cref{fig:data_distribution}. 

% For the proposed WildFake dataset, we visualize the distribution of real and fake data with different sources in ~\cref{fig:data_distribution_fake} and ~\cref{fig:data_distribution_real}, respectively. 
% In detail, for all generators in ~\cref{fig:data_distribution_fake},  $1\%5$ samples are randomly selected from fake images generated by each generator as the testing set, while the remaining fake images belong to the training set. 

% \begin{table*}[h]
% \centering
% \caption{Results of cross-generator evaluation on different training and testing subsets using ResNet-5} \label{tab:cross-generator}
% \begin{tabular}{c|c|c|c|c}
% \toprule
% \multirow{2}{*}{Training Subset}&
% \multicolumn{3}{c|}{Testing Subset}&
% \multicolumn{1}{c}{\multirow{2}{*}{Avg ACC}} \\
% \cline{2-4}
%   & Diffusion-based &GAN-based &Other-based & \multicolumn{1}{c}{}\\
% \midrule
% Diffusion-based & 99.7/99.9/99.9&  86.4/959/898 & 793/935/845 & \\

% GAN-based &771/832/743 & 981/990/996 & 914/972/946& \\

% Other-based &764/772/701 & 825/960/912 & 996/99.9/99.9 \\
% \bottomrule
% \end{tabular}
% \end{table*}
\begin{table}[htp]
\centering
\caption{Results of cross-generator evaluation on different training and testing subsets using ViT.} \label{tab:cross-generator}
\resizebox{1.0 \columnwidth}{!} {
\begin{tabular}{c|c|c|c}
\toprule
\multirow{2}{*}{Training Subset}&
\multicolumn{3}{c}{Testing Subset} \\
\cline{2-4}
  & DMs &GANs &Others \\
\midrule
DMs & \textbf{99.7/99.9/99.9} &86.4/95.9/89.8 &79.3/93.5/84.5  \\
GANs &77.1/83.2/74.3 & \textbf{98.1/99.0/99.6} & 91.4/97.2/94.6 \\
Others &76.4/77.2/70.1 & 82.5/96.0/91.2 & \textbf{99.6/99.9/99.9} \\
\bottomrule
\end{tabular}
}
\end{table}

\begin{table*}[htp]
\centering
\caption{Results of cross-architecture from DMs generator on different training and testing subsets using ViT.} \label{tab:cross-architecture}
\resizebox{2.1 \columnwidth}{!} {
\begin{tabular}{c|c|c|c|c|c|c|c|c}
\toprule
\multirow{2}{*}{Training Subset}&
\multicolumn{8}{c}{Testing Subset} \\
\cline{2-9}
  & ADM  & DALLE  & DDIM  & DDPM  & Imagen  & VQDM &  Midjourney  & SD \\
\midrule
 ADM &100/100/100 &93.3/98.9/97.8 & 78.6/90.9/84.0 & 80.5/92.0/86.1& 96.9/99.3/99.0 & 90.6/97.6/96.0 & 84.4/92.8/89.3 & 87.6/97.0/94.1  \\

DALLE & 90.0/91.8/90.6 & 99.9/99.9/99.9 & 99.7/99.9/99.9& 98.3/99.8/99.7 & 99.9/99.9/99.9 & 79.0/77.7/71.1 & 80.4/78.8/72.3 & 85.7/89.7/86.9 \\

 DDIM &89.9/91.0/87.1 &99.7/99.9/99.9/&99.9/99.9/99.9 & 99.7/99.9/99.9 &99.9/99.9/99.9&75.6/79.9/71.0 & 76.2/85.5/78.0/& 88.5/90.7/86.1 \\

 DDPM &92.0/89.9/88.5 &99.7/99.9/99.8 &99.9/99.9/99.9 & 99.8/99.9/99.9 &100/100/100& 76.0/74.1/64.9 & 77.0/75.7/67.8 & 89.1/88.7/86.1 \\

Imagen & 80.3/92.4/85.2 & 98.6/99.7/99.3& 95.0/98.5/97.1 & 94.7/98.3/96.4 & 100/100/100 & 81.8/89.1/85.4 & 84.0/89.8/86.9 & 77.3/91.3/82.7 \\

VQDM &91.8/98.1/96.7  & 88.4/96.5/91.3 & 79.7/85.4/79.6 & 71.0/84.0/67.1 & 98.1/99.4/98.6 & 99.9/99.9/99.9 & 95.7/98.7/97.4  & 93.5/98.0/95.8  \\

Midjourney  &99.9/99.9/99.9 & 99.9/99.9/99.9 & 99.6/99.9/99.9 & 99.4/99.9/99.9 & 100/100/100& 99.9/99.9/99.9 & 99.9/99.9/99.9 &  99.8/99.9/99.9 \\

SD &99.9/99.0/99.9 & 100/100/100  & 99.7/99.9/99.9 &99.3/99.9/99.9  & 100/100/100 & 99.9/99.9/100 & 99.9/100/99.9 &100/99.9/100  \\
\bottomrule
\end{tabular}
}
\end{table*}

\section{Experiments}
In this section, we conduct comprehensive experiments on WildFake to evaluate state-of-the-art AI-generated image datasets/detectors from multi-dimensional aspects. We also compare the generalization of detectors trained on different datasets.
First, we evaluate the generalization ability of the trained detectors, including cross-generator evaluation, cross-architecture evaluation, cross-weight evaluation, and cross-time evaluation in \cref{sec:generalization}.
Second, we focus on evaluating the robustness of the trained detectors by imitating propagation degradation problems in \cref{sec:robustness}.
% We briefly introduce the baseline methods (\emph{resp.,} baseline datasets) in \cref{sec:baseline_method} (\cref{sec:baseline_dataset}).

\subsection{Experimental Settings}
\noindent\textbf{Baselines}.
We select high-quality and large scale AI-generated image datasets (see \cref{sec:other_datasets} and \cref{tab:dataset_compare}) as baseline datasets, including
{\textit{DiffusionDB}}~\cite{wang2022diffusiondb},
{\textit{ArtiFact}}~\cite{rahman2023artifact},
{\textit{GenImage}}~\cite{zhu2023genimage}, and
{\textit{DiffusionForensics}}~\cite{wang2023dire}. Each of these datasets follows their original train-test split strategies.
% {\textit{CiFAKE}}~\cite{bird2023cifake},
% Considering the fact that the proposed WildFake dataset is organized with a hierarchical structure, we according the original $4:1$ train-test split to classify those subsets, according to the training and testing number in each subset, we randomly sample $1/2$ real images into each subset for training and testing.  
The baseline detectors selected for evaluation in our study include {\textit{DIRE}}~\cite{wang2023dire}, {\textit{IFDL}}~\cite{guo2023hierarchical}, {\textit{multiLID}}~\cite{lorenz2023detecting}, {\textit{LASTED}}~\cite{wu2023generalizable},
\textit{ViT}~\cite{dosoViTskiy2020image} and \textit{ResNet50}~\cite{he2016deep}. For detailed descriptions of these detectors, please refer to \cref{sec:image_detectors}.
For baseline methods, we follow the experimental setting from original paper. For naive methods, ResNet50(\emph{resp.,} ViT) is pretrained on ImageNet datasets (\emph{resp.,} LAION-5B).
All training images are resize to 224 $\times$ 224, with the Adam optimizer and Exponentially Decay scheduler with an initial learning rate of $1e-4$, and batch size (\emph{resp.,} epoch) is set as $1024$ (\emph{resp.,} 15).

\noindent\textbf{Evaluation Metrics}.
Following previous AI-generated image detection methods~\cite{wang2020cnn,wang2019detecting}, we report accuracy (ACC) and average precision (AP) in our experiments to evaluate the detectors. The threshold for computing accuracy is set to 0.5 following~\cite{wang2020cnn}.
Besides, we include the Area Under the ROC Curve (AUC) as another critical metric.

\subsection{Superiority of WildFake to Baselines}
Considering the rich and diverse content of WildFake, including wild, diverse, and hierarchical-quality fake images generated by GANs generators, DMs generators, and Others generators, detectors trained on WildFake can achieve better performance compared with other existing baseline datasets. 
% We use two naive methods consisting of ResNet50 and ViT to train detectors on the proposed WildFake dataset and comparable ArtiFact dataset
For a comparative analysis, we have chosen the ArtiFact~\cite{rahman2023artifact} dataset as a baseline, considering its volume and diversity. We train detectors using the basic ResNet50~\cite{he2016deep} (\emph{resp.,} ViT~\cite{radford2021learning}) on ArtiFact~\cite{rahman2023artifact} and WildFake. Then, evaluate the both trained detectors on the testing set of all baseline datasets and WildFake. 
The comparison results are reported in \cref{tab:evaluation_superority}, we can see that detector ResNet50-WildFake (\emph{resp.,} ViT-WildFake) outperform ResNet50-ArtiFact  (\emph{resp.,} ViT-ArtiFact) over DiffusionForensics, DiffusionDB, and GenImage datasets. 
This is especially evident for DMs-centric datasets like DiffusionForensics and DiffusionDB.
% , the cross-dataset validation results are comparable to in-dataset results. 
More comparison experiments on WildFake using different detectors are provided in Supplementary.
% ResNet50-WildFake achieves better performance on GenImage compared with ResNet50-ArtiFact.
% Considering the fact that CiFAKE dataset producing low-resolution fake images based on CIFAR-10 dataset, such fake images far from the quality of current AI-Generated images. The poor performance on CiFAKE dataset indicates that the model trained on WildFake or ArtiFactS dataset have poor robustness to low-resolution fake images.

\subsection{Generalization Capability Evaluation} \label{sec:generalization}
Owing to the hierarchical structure of WildFake, the trained detectors' generalization capabilities can be evaluated across various hierarchical dimensions, a feature that is not applicable to other baseline datasets. 
As depicted in \cref{fig:framework_dataset}, WildFake is systematically divided into five distinct levels: the first level is cross-generators consisting of three types of generators, the second level is cross-architecture in DMs generators consisting of eight types of DMs architectures, the third level is the cross-weight in SD consisting of three types of weights, the last two levels are cross-version and cross-time in three types of generators consisting of typical (\emph{resp.,} early) generators and advanced (\emph{resp.,} latest) generators. 
We utilize the baseline detector ViT to conduct comprehensive generalization experiments: (1) Cross-generator experimental comparison is designed to evaluate the gap among different generators in \cref{tab:cross-generator}. (2) Comparison among cross architectures is designed to evaluate the effects of DMs generators with different architectures in \cref{tab:cross-architecture}. 
(3) Cross-weight evaluation of SD, cross-time evaluation of GANs (\emph{resp.,} Others), and cross-version evaluation of Midjourney (\emph{resp.,} SD) are reported in Supplementary due to space limit.

\begin{table*}[htp]
\centering
\caption{Robustness evaluation of different detectors trained on the proposed WildFake dataset. ``Trans'' denotes Transformation.} \label{tab:robusness}
\resizebox{2.05\columnwidth}{!} {
\begin{tabular}{c|c|c|c|c|c|c|c|c|c}
\toprule
\multirow{2}{*}{Method}&
\multicolumn{2}{c|}{DownSample}&
\multicolumn{2}{c|}{Compression}&
\multicolumn{2}{c|}{Geometric Transformation}&
\multicolumn{2}{c|}{Watemarks}&
\multicolumn{1}{c}{\multirow{2}{*}{Color Trans}} \\
\cline{2-9}
  &128 &64 &q=70 &q=35 &Flip &Crop &Text&Image &\multicolumn{1}{c}{} \\
\midrule
% DIRE &&&&&&&&&\\

% IFDL &&&&&&&&&\\

% Multi-LID &&&&&&&&&\\

% LASTED &&&&&&&&&\\

ResNet50 &91.1/95.5/93.1 &71.3/65.0/39.8 & 84.6/95.9/92.5 &85/93.1/87.7   &95.1/98.4/96.4&91.3/98.7/96.9  & 91.0/98.8/94.1 &90.8/98.8/93.9&87.9/97.3/94.8\\

ViT &91.8/94.6/92.4&79.3/78.2/66.2 &92.4/98.1/96.0 &86.6/95.1/95 &97.1/99.8/99.4&98.9/99.9/99.1 &93.6/99.3/96.6&92.9/99.3/96.5& 98.5/99.9/99.8\\

\bottomrule
\end{tabular}
}
\end{table*}

\noindent\textbf{Evaluation on Cross-Generator}. \label{sec:cross-generator}
We first assess the performance of the ViT detector when trained and tested on images generated by the same type of generator within WildFake. 
Our WildFake dataset consists of three types of generators including DMs, GANs, and Others. Accordingly, we divide the WildFake dataset into three subsets, each with its training and testing set, based on the generator type. We then train the ViT model separately on each generator type and evaluate its performance on the corresponding testing set of each type.
The results in \cref{tab:cross-generator} indicate that the in-domain generalization ability is significantly superior to cross-domain generalization. Notably, models trained on DMs exhibit a lesser degree of generalization compared to those trained on GANs and Others. This suggests that the disparity between DMs and GANs (\emph{resp}., Others) is more pronounced than that between GANs and Others. 
We hypothesize that this is due to the distinct image generation approaches: while Others and GANs typically employ one-step inference for image generation, DMs utilize multiple denoising steps.
% Within each type generator, we further divide the data into training and testing sets according to the ratio of 4:1.

% \noindent\textbf{DMs Generators}
% \noindent\textbf{GANs Generators}
% \noindent\textbf{Other-type Generators}

\noindent\textbf{Evaluation on Cross-Architecture of DMs} \label{sec:cross-achitecture}
Considering the high quality of images generated by DMs, we further classify images from DMs into 8 categories according to the difference of architectures consisting of SD, DDPM, DDIM, ADM, DALLE, Imagen, Midjourney, and VQDM.
% In \cref{tab:cross-weight}, we can see that in-architecture testing performance reach to high level.
The performance of detectors over cross-architecture scenarios is observed to be worse to that of in-architecture validations. This suggests that different architectures within DMs generators might produce fake images with varying levels of sophistication. Another notable finding is that models trained on Midjourney and SD demonstrate superior generalization ability compared to other architectures. The reasons for this are threefold: (1) The volumn of training images from Midjourney and SD  s greater than that of other architectures, offering a more extensive learning base. (2) The content diversity of fake images from Midjourney and SD is richer, providing a broader spectrum of data for model training. (3) A portion of the fake images in Midjourney and SD are sourced from open community platforms, typically exhibiting higher quality compared to those from other architectures.

% \noindent{Cross-time evaluation on Other-based Generators}

% \noindent{Cross-time evaluation on GANs Generators}

% \noindent{Cross-time evaluation on DMs Generators}

% \subsubsection{Evaluation over Cross-dataset}
% To compare the performance of the baseline method on different dataset, we design cross-dataset experiment
% In table \cref{tab:cross-dataset}

\subsection{Robustness to Degraded Images} \label{sec:robustness} 
Images often suffer from degradation problems during propagation~\cite{schettini2010underwater}, such as low resolution, noise interference, and watermarks. It is crucial for detectors to demonstrate resilience against these challenges. To assess the robustness of detectors trained on the WildFake dataset in handling degraded images, we implement a series of degradation techniques on the testing set images:
(1) DownSample: down-sampling the high-resolution images to resolutions of 128 or 64. 
(2) Compression: introducing compression artifacts to the testing set by applying JPEG compression with quality ratios on the original test images. 
(3) Geometric Transformation: Randomly flipping or cropping the images from the testing set. 
(4) Watermark: randomly adding textual or visual watermarks on the random position of images from the testing set. (5) Color Transformation: we randomly change the brightness, contrast, saturation, and hue of images from the testing set. 
We report the robustness results of ResNet50-WildFake and ViT-WildFake over those degraded testing images in \cref{tab:robusness}. Other baseline methods evaluated with this degradation settings on WildFake are provided in Supplementary. 
Analysis of the results reveals that the ViT-based detector exhibits superior performance on these degraded images compared to the ResNet-50-based detector. 
The ResNet-50-based detector shows greater sensitivity to geometric and color transformations, whereas the ViT-based detector demonstrates better robustness. Both detectors experience a decline in performance when confronted with image-based or text-based watermarks
It is also observed that lower resolutions (\emph{resp.,} low quality) significantly affect the accuracy of the trained detectors.

% To address this, we propose evaluating the performance of the detector on these degraded images, which more accurately simulate practical conditions, as shown in \cref{tab:robusness}.
% After training the detector on our WildFake dataset, we employ a series of methods to degrade only the testing set images.
% In detail, 
% For visual watermarks, we randomly selected logo from LOGO2K dataset~\cite{wang2020logo}. 

\section{Conclusion}
In this paper, we present a large-scale dataset WildFake,  to assess the generalizability and robustness of fake image detection.
WildFake amasses a diverse array of fake images from the open-source community, featuring a wide range of image classes and styles. The dataset includes fake images generated by various types of generators, encompassing GANs, diffusion models, and other generative models.
The key strengths of WildFake notably enhance the generalization and robustness of detectors trained with this dataset, showcasing its significant applicability and effectiveness in real-world scenarios for AI-generated image detection. 
Furthermore, our in-depth evaluation experiments are designed to provide substantial insights into the capabilities of generative models at different levels, a unique benefit derived from WildFake's distinct hierarchical structure.

\appendix
\section*{\Large Appendix}

\begin{table*}[htp]
\centering
\caption{Cross-dataset generalization results of DIRE detector trained on different datasets.} \label{tab:cross-dataset-dire}
\resizebox{2 \columnwidth}{!} {
\begin{tabular}{c|c|c|c|c|c|c}
\toprule
\multirow{2}{*}{Training Dataset}&
\multicolumn{5}{c|}{Testing Dataset}&
\multicolumn{1}{c}{\multirow{2}{*}{Avg}} \\
\cline{2-6}
  & DiffusionForensics  & GenImage & DiffusionDB & ArtiFact & WildFake & \multicolumn{1}{c}{} \\
\midrule

GenImage& 82.9/93.9/76.7&99.5/99.9/99.9&91.3/94.5/92.5&59.1/61.8/52.7&72.1/92.3/80.4& 80.9/88.4/80.4 \\

DiffusionDB&82.7/82.7/53.4 &50.0/49.3/47.1&99.9/99.9/99.9&60.3/60.3/50.1&79.1/81.0/72.0& 74.4/74.6/64.5 \\

ArtiFact&79.5/88.3/74.6&75.4/82.0/82.3&59.9/64.8/62.5&92.4/92.6/92.4&69.8/83.4/70.2&75.4/82.2/76.4 \\

WildFake&85.5/97.5/84.9& 77.3/85.1/84.5&97.2/99.3/99.3&67.6/84.0/74.8 &89.3/89.6/89.7 &\bf{83.4/91.1/86.6} \\

\bottomrule
\end{tabular}
}
\end{table*}
\begin{table*}[htp]
\centering
\caption{Cross-dataset generalization results of IFDL detector trained on different datasets.} \label{tab:cross-dataset-IFDL}
\resizebox{2 \columnwidth}{!} {
\begin{tabular}{c|c|c|c|c|c|c}
\toprule
\multirow{2}{*}{Training Dataset}&
\multicolumn{5}{c|}{Testing Dataset}&
\multicolumn{1}{c}{\multirow{2}{*}{Avg}} \\
\cline{2-6}
  & DiffusionForensics  & GenImage & DiffusionDB & ArtiFact & WildFake & \multicolumn{1}{c}{} \\
\midrule

GenImage&86.4/95.9/81.4 &99.6/99.9/99.9&93.7/96.8/95.0&61.4/67.4/55.1&74.2/93.7/83.9& 83.0/90.7/83.0  \\

DiffusionDB& 88.3/88.1/53.9&52.9/51.8/51.0&99.9/100/100&63.4/62.4/53.5&85.5/85.9/76.5&78.0/77.6/66.9  \\

ArtiFact&87.9/95.8/80.0 &78.3/87.1/84.9&66.2/72.1/69.1&97.6/99.6/99.6&71.9/90.0/75.5&80.3/88.9/81.8  \\

WildFake& 88.6/97.9/92.8 &85.1/95.0/89.9& 97.7/99.5/99.7&67.7/82.1/72.4& 99.3/99.9/99.9 & \bf{87.6/93.9/99.0} \\

\bottomrule
\end{tabular}
}
\end{table*}
\begin{table*}[ht]
\centering
\caption{Cross-dataset generalization results of Multi-LID detector trained on different datasets.} \label{tab:cross-dataset-Multi-LID}
\resizebox{2 \columnwidth}{!} {
\begin{tabular}{c|c|c|c|c|c|c}
\toprule
\multirow{2}{*}{Training Dataset}&
\multicolumn{5}{c|}{Testing Dataset}&
\multicolumn{1}{c}{\multirow{2}{*}{Avg}} \\
\cline{2-6}
  & DiffusionForensics  & GenImage & DiffusionDB & ArtiFact & WildFake & \multicolumn{1}{c}{} \\
\midrule

GenImage& 58.8/59.9/51.5&75.8/73.9/74.9&59.1/60.4/52.0&49.9/45.3/42.9&50.5/61.5/54.3& 58.8/60.2/55.1  \\

DiffusionDB&50.0/50.8/45.6 &47.0/46.5/44.9&77.2/72.4/73.4&49.9/51.9/49.0&51.0/49.4/48.9& 55.0/54.2/52.3  \\

ArtiFact&55.7/61.4/54.7 & 53.6/59.6/58.4& 49.9/50.9/49.7&71.0/72.7/73.5& 50.8/59.7/51.9&56.2/60.9/57.4  \\

WildFake &56.7/58.8/54.4&52.1/58.2/58.1& 62.1/64.5/64.6&51.2/55.1/56.0& 74.3/75.9/75.9&\bf{59.2/62.5/61.8}  \\

\bottomrule
\end{tabular}
}
\end{table*}
\begin{table*}[htp]
\centering
\caption{Cross-dataset generalization results of LASTED detector trained on different datasets.} \label{tab:cross-dataset-LASTED}
\resizebox{2 \columnwidth}{!} {
\begin{tabular}{c|c|c|c|c|c|c}
\toprule
\multirow{2}{*}{Training Dataset}&
\multicolumn{5}{c|}{Testing Dataset}&
\multicolumn{1}{c}{\multirow{2}{*}{Avg}} \\
\cline{2-6}
  & DiffusionForensics  & GenImage & DiffusionDB & ArtiFact & WildFake & \multicolumn{1}{c}{} \\
\midrule

GenImage &87.6/96.2/83.7 &99.8/99.9/99.9&94.3/97.9/95.7&63.4/66.7/57.0&78.8/94.3/86.9& {84.7/91.0/84.6}\\

DiffusionDB &90.7/90.5/58.4 &53.1/53.2/52.3&99.9/100/100&65.4/66.5/55.9&88.2/90.0/79.1&{79.4/80.0/69.1} \\

ArtiFact& 91.3/98.6/84.1& 80.2/89.1/89.2&69.5/75.7/72.6&98.2/99.7/99.6& 73.0/92.0/77.8&82.4/91.1/84.7 \\

WildFake & 94.9/98.1/96.2 &87.0/91.4/93.8 &98.9/99.7/99.8 &71.4/89.0/79.1 & 99.7/99.9/99.9& \bf{93.0/95.6/93.7}\\

\bottomrule
\end{tabular}
}
\end{table*}
\begin{table*}[htp]
\centering
\caption{Cross-dataset generalization results of ResNet50 detector trained on different datasets.} \label{tab:cross-dataset-resnet50}
\resizebox{2 \columnwidth}{!} {
\begin{tabular}{c|c|c|c|c|c|c}
\toprule
\multirow{2}{*}{Training Dataset}&
\multicolumn{5}{c|}{Testing Dataset}&
\multicolumn{1}{c}{\multirow{2}{*}{Avg}} \\
\cline{2-6}
  & DiffusionForensics  & GenImage & DiffusionDB & ArtiFact & WildFake & \multicolumn{1}{c}{} \\
\midrule

GenImage&84.2/95.6/77.9 &99.7/99.9/99.9&93.0/96.3/94.3&61.3/66.1/52.8 &71.6/92.1/79.0&81.9/90.0/80.7 \\

DiffusionDB& 84.2/84.1/49.4& 50.1/49.3/48.5&99.9/100/100&61.3/61.4/50.0&81.1/82.7/72.8&75.3/75.5/64.1 \\

ArtiFact& 85.4/94.9/76.7  &76.5/84.8/82.9 & 64.1/69.9/68.1 &{97.2/99.5/99.3}& 85.4/94.9/76.7 & 81.7/88.8/87.4  \\

WildFake &{87.2/96.6/83.4}  & {89.0/89.9/89.3} &{96.3/99.2/99.2} &{68.0/84.7/75.3} & {99.6/99.9/99.9} & \bf{86.4/94.1/89.4}  \\

\bottomrule
\end{tabular}
}
\end{table*}
\begin{table*}[htp]
\centering
\caption{Cross-dataset generalization results of ViT detector trained on different datasets.} \label{tab:cross-dataset-vit}
\resizebox{2 \columnwidth}{!} {
\begin{tabular}{c|c|c|c|c|c|c}
\toprule
\multirow{2}{*}{Training Dataset}&
\multicolumn{5}{c|}{Testing Dataset}&
\multicolumn{1}{c}{\multirow{2}{*}{Avg}} \\
\cline{2-6}
  & DiffusionForensics  & GenImage & DiffusionDB & ArtiFact & WildFake & \multicolumn{1}{c}{} \\
\midrule

GenImage& 84.2/97.2/85.3&99.6/99.9/99.9&97.2/99.0/98.6&61.3/61.1/49.8&76.8/93.8/83.3&83.8/90.2/83.4 \\

DiffusionDB & 84.2/83.6/47.8&50.0/46.4/42.8&99.9/100/100&61.2/61.5/50.2&80.4/81.9/71.4&75.2/74.6/62.4 \\

ArtiFact& 84.2/96.1/82.5  &78.5/88.1/85.0 & 68.4/75.3/73.2 &{96.8/99.6/99.5} & 84.2/96.1/82.5  & {82.4/91.0/84.4}\\

WildFake& {95.8/99.1/97.2}  & {88.6/83.6/89.7} &{99.3/99.8/99.9} & 62.2/81.9/68.8 &{99.1/99.9/99.9} &\bf{89.0/92.8/91.1} \\

\bottomrule
\end{tabular}
}
\end{table*}

\begin{table*}[htp]
\centering
\caption{Cross-dataset generalization comparison of different detectors trained on GenImage dataset.} \label{tab:cross-dataset-genimage}
\resizebox{2 \columnwidth}{!} {
\begin{tabular}{c|c|c|c|c|c|c}
\toprule
\multirow{2}{*}{Detectors}&
\multicolumn{5}{c|}{Testing Dataset}&
\multicolumn{1}{c}{\multirow{2}{*}{Avg}} \\
\cline{2-6}
  & DiffusionForensics  & GenImage & DiffusionDB & ArtiFact & WildFake & \multicolumn{1}{c}{} \\
\midrule

DIRE & 82.9/93.9/76.7&99.5/99.9/99.9&91.3/94.5/92.5&59.1/61.8/52.7&72.1/92.3/80.4& 80.9/88.4/80.4 \\

IFDL  &86.4/95.9/81.4 &99.6/99.9/99.9&93.7/96.8/95.0&61.4/67.4/55.1&74.2/93.7/83.9& 83.0/90.7/83.0  \\

Multi-LID & 58.8/59.9/51.5&75.8/73.9/74.9&59.1/60.4/52.0&49.9/45.3/42.9&50.5/61.5/54.3& 58.8/60.2/55.1  \\

LASTED  &87.6/96.2/83.7 &99.8/99.9/99.9&94.3/97.9/95.7&63.4/66.7/57.0&78.8/94.3/86.9& \bf{84.7/91.0/84.6}\\

ResNet50&84.2/95.6/77.9 &99.7/99.9/99.9&93.0/96.3/94.3&61.3/66.1/52.8 &71.6/92.1/79.0&81.9/90.0/87.0 \\
ViT & 84.2/97.2/85.3&99.6/99.9/99.9&97.2/99.0/98.6&61.3/61.1/49.8&76.8/93.8/83.3&{83.8/92.0/83.4}\\

\bottomrule
\end{tabular}
}
\end{table*}
\begin{table*}[htp]
\centering
\caption{Cross-dataset generalization comparison of different detectors trained on DiffusionDB dataset.} \label{tab:cross-dataset-diffusiondb}
\resizebox{2 \columnwidth}{!} {
\begin{tabular}{c|c|c|c|c|c|c}
\toprule
\multirow{2}{*}{Detectors}&
\multicolumn{5}{c|}{Testing Dataset}&
\multicolumn{1}{c}{\multirow{2}{*}{Avg}} \\
\cline{2-6}
  & DiffusionForensics  & GenImage & DiffusionDB & ArtiFact & WildFake & \multicolumn{1}{c}{} \\
\midrule

DIRE &82.7/82.7/53.4 &50.0/49.3/47.1&99.9/99.9/99.9&60.3/60.3/50.1&79.1/81.0/72.0& 74.4/74.6/64.5 \\

IFDL  & 88.3/88.1/53.9&52.9/51.8/51.0&99.9/100/100&63.4/62.4/53.5&85.5/85.9/76.5&78.0/77.6/66.9  \\

Multi-LID &50.0/50.8/45.6 &47.0/46.5/44.9&77.2/72.4/73.4&49.9/51.9/49.0&51.0/49.4/48.9& 55.0/54.2/52.3  \\

LASTED  &90.7/90.5/58.4 &53.1/53.2/52.3&99.9/100/100&65.4/66.5/55.9&88.2/90.0/79.1& \bf{79.4/80.0/69.1} \\

ResNet50& 84.2/84.1/49.4& 51.0/49.3/48.5&99.9/100/100&61.3/61.4/50.0&81.1/82.7/72.8&75.3/75.5/64.1 \\
ViT & 84.2/83.6/47.8&50.0/46.4/42.8&99.9/100/100&61.2/61.5/52.0&84.0/81.9/71.4&75.2/74.6/62.4 \\

\bottomrule
\end{tabular}
}
\end{table*}
\begin{table*}[htp]
\centering
\caption{Cross-dataset generalization comparison of different detectors trained on ArtiFact dataset.} \label{tab:cross-dataset-artifact}
\resizebox{2 \columnwidth}{!} {
\begin{tabular}{c|c|c|c|c|c|c}
\toprule
\multirow{2}{*}{Detectors}&
\multicolumn{5}{c|}{Testing Dataset}&
\multicolumn{1}{c}{\multirow{2}{*}{Avg}} \\
\cline{2-6}
  & DiffusionForensics  & GenImage & DiffusionDB & ArtiFact & WildFake & \multicolumn{1}{c}{} \\
\midrule

DIRE &79.5/88.3/74.6&75.4/82.0/82.3&59.9/64.8/62.5&92.4/92.6/92.4&69.8/83.4/70.2&75.4/82.2/76.4 \\

IFDL  &87.9/95.8/80.0 &78.3/87.1/84.9&66.2/72.1/69.1&97.6/99.6/99.6&71.9/90.0/75.5&80.3/88.9/81.8  \\

Multi-LID &55.7/61.4/54.7 & 53.6/59.6/58.4& 49.9/50.9/49.7&71.0/72.7/73.5& 50.8/59.7/51.9&56.2/60.9/57.4  \\

LASTED  & 91.3/98.6/84.1& 80.2/89.1/89.2&69.5/75.7/72.6&98.2/99.7/99.6& 73.0/92.0/77.8&\bf{82.4/91.1/84.7} \\

ResNet50& 85.4/94.9/76.7  &76.5/84.8/82.9 & 64.1/69.9/68.1 &{97.2/99.5/99.3}& 85.4/94.9/76.7 & 81.7/88.8/87.4  \\
ViT & 84.2/96.1/82.5  &78.5/88.1/85.0 & 68.4/75.3/73.2 &{96.8/99.6/99.5} & 84.2/96.1/82.5  & {82.4/91.0/84.4}\\

\bottomrule
\end{tabular}
}
\end{table*}
\begin{table*}[htp]
\centering
\caption{Cross-dataset generalization comparison of different detectors trained on the proposed WildFake dataset.} \label{tab:cross-dataset}
\resizebox{2 \columnwidth}{!} {
\begin{tabular}{c|c|c|c|c|c|c}
\toprule
\multirow{2}{*}{Detectors}&
\multicolumn{5}{c|}{Testing Dataset}&
\multicolumn{1}{c}{\multirow{2}{*}{Avg}} \\
\cline{2-6}
  & DiffusionForensics  & GenImage & DiffusionDB & ArtiFact & WildFake & \multicolumn{1}{c}{} \\
\midrule

DIRE &85.5/97.5/84.9& 77.3/85.1/84.5&97.2/99.3/99.3&67.6/84.0/74.8 &89.3/89.6/89.7 &83.4/91.1/86.6 \\

IFDL & 88.6/97.9/92.8 &85.1/95.0/89.9& 97.7/99.5/99.7&67.7/82.1/72.4& 99.3/99.9/99.9 & 87.6/93.9/99.0 \\

Multi-LID &56.7/58.8/54.4&52.1/58.2/58.1& 62.1/64.5/64.6&51.2/55.1/56.0& 74.3/75.9/75.9&59.2/62.5/61.8  \\

LASTED  & 94.9/98.1/96.2 &87.0/91.4/93.8 &98.9/99.7/99.8 &71.4/89.0/79.1 & 99.7/99.9/99.9& \bf{93.0/95.6/93.7}\\

ResNet50  &{87.2/96.6/83.4}  & {89.0/89.9/89.3} &{96.3/99.2/99.2} &68.0/84.7/75.3 & {99.6/99.9/99.9} & {86.4/94.1/89.4} \\

ViT & {95.8/99.1/97.2}  & {88.6/83.6/89.7} &{99.3/99.8/99.9} & 62.2/81.9/68.8 &{99.1/99.9/99.9} &{89.0/92.8/91.1} \\

\bottomrule
\end{tabular}
}
\end{table*}

% \begin{table*}[h]
% \centering
% \caption{Results of cross-weight from Stable Diffusion evaluation on different training and testing subsets using ViT. ACC($\%$), AP($\%$), and AUC($\%$) are reported. (ACC/AP/AUC in the Table).} \label{tab:cross-weight}
% \begin{tabular}{c|c|c|c|c}
% \toprule
% \multirow{2}{*}{Training Subset}&
% \multicolumn{3}{c|}{Testing Subset}&
% \multicolumn{1}{c}{\multirow{2}{*}{Avg ACC}} \\
% \cline{2-4}
%   &Original SD &Personalized SD &SD with Adaptor &\multicolumn{1}{c}{}\\
% \midrule

% Original SD &999/1.0 /999  &998/999/999&998/999/999 &\\

% Personalized SD &997/999/998&999/999/999&998/999/999 &\\

% SD with Adaptor &999/999/999 &999/999/999 &999/999/999 &\\
% \bottomrule
% \end{tabular}
% \end{table*}

\begin{table}[htp]
\centering
\caption{Results of cross-weight from SD evaluation on different training and testing subsets using ViT. } \label{tab:cross-weight}
\resizebox{1 \columnwidth}{!} {
\begin{tabular}{c|c|c|c}
\toprule
\multirow{2}{*}{Training Subset}&
\multicolumn{3}{c}{Testing Subset}\\
\cline{2-4}
  &Original SD &Personalized SD &SD with Adaptor\\
\midrule
Original SD &99.9/100 /99.9  &99.8/99.9/99.9&99.8/99.9/99.9 \\
Personalized SD &99.7/99.9/99.8&99.9/99.9/99.9&99.8/99.9/99.9 \\
SD with Adaptor &99.9/99.9/99.9 &99.9/99.9/99.9 &99.9/99.9/99.9 \\
\bottomrule
\end{tabular}
}
\end{table}

\begin{table}[htp]
\centering
\caption{Results of cross-version evaluation over SD from DMs generators using ViT.} \label{tab:cross-time-diffusion-1}
\resizebox{0.8 \columnwidth}{!} {
\begin{tabular}{c|c|c}
\toprule
\multirow{2}{*}{Training Subset}&
\multicolumn{2}{c}{Testing Subset} \\
\cline{2-3}
&Advanced &Typical\\
\midrule
Advanced&100 /100 /99.9&99.5/99.9/99.9    \\
\midrule
Typical&99.9/99.9/99.9&99.9/99.9/99.9  \\

\bottomrule
\end{tabular}
}
\end{table}

\begin{table}[htp]
\centering
\caption{Results of cross-version evaluation over Midjourney from DMs generators using ViT.} \label{tab:cross-time-diffusion-2}
\resizebox{0.8 \columnwidth}{!} {
\begin{tabular}{c|c|c}
\toprule
\multirow{2}{*}{Training Subset}&
\multicolumn{2}{c}{Testing Subset}\\
\cline{2-3}
&Advanced &Typical\\
\midrule
Advanced&99.5/99.9/99.9&99.5/99.9/99.9   \\
\midrule
Typical&99.9/99.9/99.9&99.9/99.9/100    \\

\bottomrule
\end{tabular}
}
\end{table}

\begin{table}[htp]
\centering
\caption{Results of cross-time evaluation over GANs generators using ViT.} \label{tab:cross-time-gan}
\resizebox{0.8 \columnwidth}{!} {
\begin{tabular}{c|c|c}
\toprule
\multirow{2}{*}{Training Subset}&
\multicolumn{2}{c}{Testing Subset} \\
\cline{2-3}
&Latest&Early \\
\midrule
Latest&99.9/100/100&81.1/39.4/64.3  \\
Early&66.6/67.8/51.1&97.5/96.6/99.2 \\
\bottomrule
\end{tabular}
}
\end{table}

% \begin{table}[htp]
% \centering
% \caption{Results of cross-time evaluation over Others generators using ViT.} \label{tab:cross-time-other}
% \resizebox{0.8 \columnwidth}{!} {
% \begin{tabular}{c|c|c}
% \toprule
% \multirow{2}{*}{Training Subset}&
% \multicolumn{2}{c}{Testing Subset} \\
% \cline{2-3}
% &Latest&Early\\
% \midrule
% Latest&99.9/99.9/99.9&75.6/42.1/53.8   \\
% Early&94.3/98.4/96.1&99.5/99.7/99.8  \\

% \bottomrule
% \end{tabular}
% }
% \end{table}

% \begin{table}[htp]
% \centering
% \caption{Results of cross-version evaluation over SD from DMs generators using ViT.} \label{tab:cross-time-diffusion-1}
% \resizebox{0.8 \columnwidth}{!} {
% \begin{tabular}{c|c|c}
% \hline
% \multirow{2}{*}{Training Subset}&
% \multicolumn{2}{c}{Testing Subset} \\
% \cline{2-3}
% &Advanced &Typical\\
% \hline
% Advanced&100 /100 /99.9&99.5/99.9/99.9    \\
% \hline
% Typical&99.9/99.9/99.9&99.9/99.9/99.9  \\

% \hline
% \end{tabular}
% }
% \end{table}

% \begin{table}[htp]
% \centering
% \caption{Results of cross-version evaluation over Midjourney from DMs generators using ViT.} \label{tab:cross-time-diffusion-2}
% \resizebox{0.8 \columnwidth}{!} {
% \begin{tabular}{c|c|c}
% \hline
% \multirow{2}{*}{Training Subset}&
% \multicolumn{2}{c}{Testing Subset}\\
% \cline{2-3}
% &Advanced &Typical\\
% \hline
% Advanced&99.5/99.9/99.9&99.5/99.9/99.9   \\
% \hline
% Typical&99.9/99.9/99.9&99.9/99.9/100    \\

% \hline
% \end{tabular}
% }
% \end{table}

\begin{table*}[htp]
\centering
\caption{Robustness evaluation of different detectors trained on the proposed WildFake dataset. ``Trans'' denotes Transformation.} \label{tab:robusness_supple}
\resizebox{2\columnwidth}{!} {
\begin{tabular}{c|c|c|c|c|c|c|c|c|c}
\toprule
\multirow{2}{*}{Method}&
\multicolumn{2}{c|}{DownSample}&
\multicolumn{2}{c|}{Compression}&
\multicolumn{2}{c|}{Geometric Transformation}&
\multicolumn{2}{c|}{Watemarks}&
\multicolumn{1}{c}{\multirow{2}{*}{Color Trans}} \\
\cline{2-9}
 &128&64&q=70&q=35&Flip&Crop&Text&Image&\multicolumn{1}{c}{} \\
\midrule
DIRE&85.8/88.9/87.6&61.6/56.3/33.9&75.6/82.7/81.0&65.4/75.5/71.2&85.0/91.8/89.3&86.3/92.5/91.6&87.7/94.2/90.6&86.5/93.8/89.4&81.3/89.1/86.8\\

IFDL&91.3/95.2/93.4&73.1/69.6/49.4&87.4/97.8/95.1&82.9/94.1/89.3&94.1/99.4/98.2&94.9/99.6/99.1&93.8/99.3/96.9&91.3/99.1/94.9&96.0/98.2/97.1\\

Multi-LID&59.3/62.4/61.5&51.3/50.2/35.8&55.7/58.6/57.5&53.3/56.8/54.6&58.9/64.1/61.6&59.5/64.4/63.3&60.0/61.2/59.1&58.0/62.2/59.0&58.3/59.4/61.6\\

LASTED&\bf{92.5/96.5/93.2}&75.8/71.3/51.8&89.6/97.7/95.0&\bf{88.5/95.7/91.5}&96.1/99.3/99.0&98.0/99.6/99.6&\bf{95.4/99.4/97.6}&\bf{94.7/99.5/97.4}&92.3/99.8/99.6\\

ResNet50&91.1/95.5/93.1&71.3/65.0/39.8&84.6/95.9/92.5&85.0/93.1/87.7  &95.1/98.4/96.4&91.3/98.7/96.9 &91.0/98.8/94.1&90.8/98.8/93.9&87.9/97.3/94.8\\

ViT&91.8/94.6/92.4&\bf{79.3/78.2/66.2}&\bf{92.4/98.1/96.0}&86.6/95.1/95.0&\bf{97.1/99.8/99.4}&\bf{98.9/99.9/99.1}&93.6/99.3/96.6&92.9/99.3/96.5&\bf{98.5/99.9/99.8}\\

\bottomrule
\end{tabular}
}
\end{table*}

In this supplementary material, we provide detailed information and additional analyses that complements our main submission. The sections are structured as follows: In \cref{sec:implementary_details}, we provide a comprehensive overview of the baseline datasets and detectors utilized in our study, providing a foundation for understanding the experimental setup and detector methodologies. In \cref{sec:superior}, we present more cross-dataset validation comparisons. This section aims to underscore the advancements and superiority of our proposed WildFake dataset in comparison to existing benchmarks. In \cref{sec:compare_methods}, we systematically compare the performance of various baseline detectors across different datasets. This comparative analysis is pivotal in evaluating the effectiveness and adaptability of each detector. In \cref{sec:generalization}, we present cross-weight, cross-time, and cross-version evaluations. It offers an in-depth analysis of the proposed WildFake dataset, highlighting its robustness and versatility in various scenarios. Finally, in \cref{sec:robustness_supple}, we examine and compare the robustness of different detectors when trained on the proposed WildFake dataset, aiming to showcase the resilience of these detectors against various challenges and degradation scenarios.

\section{Experimental Details} \label{sec:implementary_details}
\noindent\textbf{Baseline Detectors}.
\begin{itemize}[leftmargin=*]
\item {\textit{DIRE}}~\cite{wang2023dire} is specifically designed to differentiate between images generated by DMs and real images. It achieves this by measuring the discrepancy between an input image and its reconstructed counterpart using a pre-trained DM. 
In accordance with the original setup of {DIRE}~\cite{wang2023dire}, we utilize the ADM network\cite{dhariwal2021diffusion}, pre-trained on the LSUN-Bedroom dataset~\cite{yu2015lsun}, as the reconstruction model. 
This model is employed to calculate reconstructed images for each input. For obtaining these reconstructed images, we adopt the DDIM~\cite{song2020denoising} inversion and reconstruction process, with the number of diffusion steps set to 20 by default. A ResNet-50 model~\cite{he2016deep} is used as the classifier in this framework.

\item {\textit{IFDL}}~\cite{guo2023hierarchical} leverages multi-branch feature extractor, along with localization and classification modules, designed to represent the attributes of a fake image with multiple labels at different levels. 
In line with the original setup in IFDL~\cite{guo2023hierarchical}, the IFDL model is trained for 400,000 iterations with a batch size of 16, evenly split between 8 real and 8 fake images.
Given that images in our WildFake dataset are categorized into five-level subsets, including cross-generator, cross-architecture, cross-weight, cross-version, and cross-time, we have adapted the IFDL~\cite{guo2023hierarchical} method accordingly. This adaptation involves a modification based on these five levels, setting all masks to $0$, as our current focus is solely on fully-synthesized images. 

\item {\textit{Multi-LID}}~\cite{lorenz2023detecting}  focuses on extracting feature-map representations of fake images and calculating multi-local intrinsic dimensionality to train a classifier. 
Following the settings in {Multi-LID}~\cite{lorenz2023detecting},the process begins by calculating the standard mean and deviation across the dataset to normalize the inputs, ensuring a uniform data distribution. 
Then, images are processed through an untrained ResNet18 model\cite{he2016deep} to extract their features, which are then utilized to compute Multi-LID scores. The experimental setup includes a training dataset comprising 1,600 samples per class. 
Finally, a random forest classifier is trained using the labeled Multi-LID scores to discern between real and fake images.

\item {\textit{LASTED}}~\cite{wu2023generalizable} treats synthetic image detection as an identification problem, enhancing training images with meticulously designed textual labels for joint image-text contrastive learning.
The implementation of {{LASTED}}~\cite{wu2023generalizable} employs the Adam optimizer~\cite{kingma2014adam}, with an initial learning rate set at $1e-4$. The learning rate is halved if validation accuracy does not improve for two consecutive epochs, continuing until convergence is achieved. 
During the training and testing phases, all input images are randomly or center cropped into $448 \times 448$ patches. Image augmentation techniques, such as compression, blurring, and scaling, are applied with a $50\%$ probability. The batch size for the process is set at $48$. The batch size is set to 48. 
Given that images from the WildFake dataset and other baseline datasets do not come with specific style labels, we adhere to the $\mathcal{R}_1$ setting in  {LASTED}~\cite{wu2023generalizable}.
Here, textual labels are defined simply as ``Real, Synthetic'' during the training stage.

\item \textit{Naive methods} utilize pre-trained discriminative models, specifically \textit{ViT}\cite{dosoViTskiy2020image} and \textit{ResNet50}\cite{he2016deep}, to differentiate between generated and real images. 
For this purpose, the ResNet50 model (\emph{resp.,} ViT) is pre-trained on the ImageNet datasets (\emph{resp.,} LAION-5B).
All training images are resized to $224 \times 224$ pixels. We employ the Adam optimizer alongside an Exponentially Decay scheduler for learning rate adjustment, starting with an initial rate of $1e-4$. The batch size is set at $1024$, with the training process lasting $15$ epochs.

\end{itemize}
\noindent\textbf{Baseline Datasets} \label{sec:baseline_dataset}
% \noindent{\textbf{CNNSpot}}~\cite{wang2020cnn} are built upon generative adversarial network, CNNSpot only employs ProGAN~\cite{karras2017progressive} for generating training set and evaluatess detector performance across a variety of GANs testing sets.
\begin{itemize}[leftmargin=*]
% \item {\textit{CiFAKE}}~\cite{bird2023cifake} exclusively employs Stable Diffusion V1.4~\cite{rombach2022high} for the generation of fake images. 

\item {\textit{DiffusionDB}}~\cite{wang2022diffusiondb} assembles images that have been shared on the Stable Diffusion public Discord server~\cite{website2023}.  
% However, it does not apply hierarchical splitting method to divide the whole dataset into  hierarchical subsets, leading to applicable cross-evaluation.

\item {\textit{ArtiFact}}~\cite{rahman2023artifact} generates fake images using real prompts sourced from actual datasets. 
In total, the dataset comprises 2,496,738 images, of which 964,989 are real and 1,531,749 are fake. The dataset predominantly features categories such as Human/Human Faces, Animal/Animal Faces, Vehicles, Places, and Art. It utilizes 13 GANs and 7 DMs for image generation. Notably, the number of fake images produced by GANs significantly exceeds those generated by DMs.

\item {\textit{GenImage}}~\cite{zhu2023genimage} primarily utilizes the 1000 classes from the ImageNet dataset~\cite{russakovsky2015imagenet} to generate its content.
It contains over one million pairs of real and fake images, fully incorporates all the real images available in ImageNet. For image generation, {\textit{GenImage}} leverages the 1000 distinct labels present in ImageNet. In total, the dataset consists of 2,681,167 images, divided into 1,331,167 real and 1,350,000 fake images. Of the real images, 1,281,167 are allocated for training purposes, and 50,000 are set aside for testing.

\item {\textit{DiffusionForensics}}~\cite{wang2023dire} utilizes a variety of DMs p to create fake images. These images are based on several well-known datasets: LSUN-Bedroom~\cite{yu2015lsun}, ImageNet~\cite{russakovsky2015imagenet}, and CelebA-HQ~\cite{karras2017progressive}. Each of these source datasets is divided into training and testing sets, adhering to their respective original split strategies. In total, the {{DiffusionForensics}} dataset comprises 439,020 fake images generated by DMs, alongside 92,000 real images.

\item {\textit{WildFake}} contains a total of 3,694,313 images, of which 1,013,446 are real and 2,680,867 are fake. We have divided both real and fake images into training and testing sets, maintaining a ratio of $4:1$ for each category. Specifically, for every generator used in the dataset, $20\%$ of the fake images they produce are randomly selected to constitute the testing set, while the remaining $80\%$ are allocated to the training set. This systematic approach ensures a balanced and representative distribution of images for both training and evaluation purposes.
\end{itemize}
% To prevent overfitting, various augmentations are also employed randomly, including scale-shift-rotate-shear, contrast-brightness-hue, and flips. 

\section{Demonstrating WildFake's Superiority over Baseline Datasets.} \label{sec:superior}
The WildFake dataset stands out due to its rich and diverse content. It contains a wide array of fake images characterized by their wild, varied, and hierarchically-structured quality, produced by an assortment of generators, including GANs, DMs, and Others. This diversity and depth in content enable detectors trained on the WildFake dataset to achieve superior performance when compared to those trained on other existing baseline datasets. The varied nature of the images within WildFake provides a more comprehensive training ground, thereby enhancing the detectors' ability to generalize and adapt to a broader range of synthetic images.
% In this section, we conduct comprehensive comparison over cross-dataset and cross-methods. In detail, we select GenImage, DiffusionDB, ArtiFact, and WildFake as baseline training datasets, and DIRE, IFDL, Multi-LID, LASTED, ResNet50, and ViT as baseline methods, all of those baseline methods are trained on selected datasets and evaluation over all baseline datasets. 
In this section, we select GenImage~\cite{zhu2023genimage}, DiffusionDB~\cite{wang2022diffusiondb}, and ArtiFact~\cite{rahman2023artifact} as comparable datasets for our analysis. 
We train the same set of detectors on these baseline datasets as well as the proposed WildFake dataset. 
This setup allows us to conduct cross-dataset experiments aimed at evaluating the cross-dataset generalization capabilities of detectors trained on various datasets. 
For this cross-data generalization validation, we employ a range of baseline detectors including DIRE~\cite{wang2023dire}, IFDL~\cite{guo2023hierarchical}, Multi-LID~\cite{lorenz2023detecting}, LASTED~\cite{wu2023generalizable}, ResNet50~\cite{he2016deep}, and ViT\cite{dosoViTskiy2020image} as baseline detectors to implement cross-dataset comparison. 
The cross-data generalization comparison results from DIRE~\cite{wang2023dire} (\emph{resp.,} IFDL~\cite{guo2023hierarchical}, Multi-LID~\cite{lorenz2023detecting}, LASTED~\cite{wu2023generalizable}, ResNet50~\cite{he2016deep}, and ViT\cite{dosoViTskiy2020image} ) are reported in \cref{tab:cross-dataset-dire} (\emph{resp.,} \cref{tab:cross-dataset-IFDL}, \cref{tab:cross-dataset-Multi-LID}, \cref{tab:cross-dataset-LASTED}, \cref{tab:cross-dataset-resnet50}, and \cref{tab:cross-dataset-vit}).
% From the comparison results, we can see that all of those baseline detector trained on the proposed WildFake dataset achieve best cross-dataset generalization ability, which indicates that construction of the proposed WildFake dataset benefits to train a detectors with high generalization ability to diverse datasets. Since the content in the proposed WildFake dataset is diverse and most of fake images with high quality are collected from open-source websites, which bridge the gap among those baseline datasets. Besides, we introduce hierarchical structure into the proposed WildFake dataset, which helps to construct comprehensive datasets cover different generalization methods. 
The comparison results demonstrate that baseline detectors trained on the proposed WildFake dataset exhibit the best cross-dataset generalization ability. 
This suggests that the diverse and high-quality content of the WildFake dataset, primarily sourced from open-source websites, plays a crucial role in training detectors with a high degree of generalization to various datasets. 
The inclusion of a hierarchical structure in WildFake further contributes to this by covering different generalization methods and ensuring a comprehensive dataset.
% Another observation is that the generalization ability of detector trained on GenImage~\cite{zhu2023genimage} or  ArtiFact~\cite{rahman2023artifact} achive better performance that trained on DiffusionDB~\cite{wang2022diffusiondb} dataset, since the  DiffusionDB~\cite{wang2022diffusiondb} dataset only collect fake images generated by diffusion-based generators with 1000 category labels from ImageNet dataset, leading to limited diversity. 
Another noteworthy observation is that detectors trained on the GenImage~\cite{zhu2023genimage} or ArtiFact~\cite{rahman2023artifact} datasets show better generalization performance compared to those trained on the DiffusionDB~\cite{wang2022diffusiondb} dataset. This can be attributed to the fact that DiffusionDB primarily collects fake images generated by diffusion-based generators using the 1000 category labels from the ImageNet dataset. Such a focused collection results in a dataset with limited diversity, which may constrain the generalization ability of the trained detectors.

\section{Comparison of Baseline Detectors} \label{sec:compare_methods}
In this section, we analyze and compare the cross-dataset generalization performance of various baseline detectors trained on the same dataset. 
% This comparison reorganizes the results previously discussed in \cref{sec:superior}. 
The baseline detectors included in our analysis are  DIRE~\cite{wang2023dire}, IFDL~\cite{guo2023hierarchical}, Multi-LID~\cite{lorenz2023detecting}, LASTED~\cite{wu2023generalizable}, ResNet50~\cite{he2016deep}, and ViT\cite{dosoViTskiy2020image}. The cross-dataset generalization comparison results of different baseline detectors trained on GenImage~\cite{zhu2023genimage}(\emph{resp.,} DiffusionDB~\cite{wang2022diffusiondb}, ArtiFact~\cite{rahman2023artifact}, and WildFake) dataset are reported in \cref{tab:cross-dataset-genimage}(\emph{resp.,} \cref{tab:cross-dataset-diffusiondb}, \cref{tab:cross-dataset-artifact}, and \cref{tab:cross-dataset}). 
In our comparative analysis, LASTED~\cite{wu2023generalizable} emerges as the top performer in terms of generalization across  GenImage~\cite{zhu2023genimage}, DiffusionDB~\cite{wang2022diffusiondb}, ArtiFact~\cite{rahman2023artifact}, and WildFake.  The results notably indicate that LASTED's use of language guidance significantly enhances its generalizability across different datasets. Another key observation is that IFDL~\cite{guo2023hierarchical} also achieves comparable performance, which can be attributed to its multi-branch feature extractor that is adept at learning fine-grained features.
In contrast, DIRE~\cite{wang2023dire}, while specifically designed for detecting diffusion-based fake images via a reconstruction method, shows limited generalization to other datasets. This limitation may stem from its focus on a specific type of generator and neglecting the variations among different generators.
Additionally, we observe that the naive methods, including ResNet50~\cite{he2016deep} and ViT~\cite{dosoViTskiy2020image}, also perform satisfactorily across these datasets. This can be explained by the fact that training with large-scale datasets, encompassing more than one million images, facilitates the development of a data-driven classifier adept at AI-generated image detection.

\section{Generalization Evaluation} \label{sec:generalization}
% Furthermore, we can also compare the cross-architecture performance in GANs (\emph{resp.,} Other-based generators). 
Leveraging the diverse and comprehensive nature of WildFake, we have developed a series of comparative experiments to assess the generalization capabilities of various detectors. 
Recognizing the popularity of Stable Diffusion, which includes numerous personalized and finetuned models as well as target-oriented adaptors, we have designed cross-weight comparison experiments. These experiments, detailed in \cref{tab:cross-weight}, aim to evaluate the generalization ability across different weights within Stable Diffusion.
% Besides, we can also conduct the cross-weight evaluation on GANs generators where each architecture is trained on different datasets. (4) 
For the three types of generators (Stable Diffusion, GANs, and Others), we perform evaluations that focus on both cross-time and cross-version aspects. This involves training detectors on Typical/Early (\emph{resp.,} Advanced/Latest) models and then testing on Advanced/Latest (\emph{resp.,} Typical/Early) models. The results of these evaluations are documented in \cref{tab:cross-time-diffusion-1}, \cref{tab:cross-time-diffusion-2}, and \cref{tab:cross-time-gan}.

\paragraph{Evaluation over Cross-Weight from SD.} \label{sec:cross-weight}
In open-source communities, users often employ varied training strategies to customize their models. These strategies may include finetuning parts of the base model's parameters, using the DreamBooth training strategy for optimizing the parameters of UNET or the text-encoder, or creating different adaptors like LORA, ControlNet, and LyCORIS to combine with the base model for controlled generation. To assess how well detectors generalize across different weights within the same SD architecture, we have conducted a cross-weight evaluation.
As indicated in \cref{tab:cross-weight}, the results show that the ability of detectors to generalize across different weights is on par with their performance in in-weight evaluations.

\paragraph{Evaluation over Cross-version and Cross-Time.} \label{sec:cross-time}
Due to the rapid development of DMs, open-source frameworks like SD and commercial entities like Midjourney are updated frequently. 
To assess the generalization capabilities between typical and advanced generators, we conducted a comprehensive cross-version evaluation. 
The results for SD are detailed in \cref{tab:cross-time-diffusion-1}, and those for Midjourney are in \cref{tab:cross-time-diffusion-2}. 
These cross-version findings suggest that advancements in the quality of fake images have a minimal impact on generalization ability. 
This phenomenon can be attributed to the fact that the core architecture of these updated generators remains grounded in DMs, and their core modules exhibit stability, leading to relatively consistent characteristics in the generated fake images. 
Furthermore, we have also compiled cross-time evaluation results for GANs in \cref{tab:cross-time-gan}. These results reveal that detectors trained on the latest GANs display superior generalization capabilities compared to those trained on earlier versions of GANs.
% Besides, we also report cross-time evaluation results for GANs and other generators in  \cref{tab:cross-time-gan} and in  \cref{tab:cross-time-other}.

\section{Robustness to Degraded Images} \label{sec:robustness_supple} 
Images commonly encounter degradation issues during propagation, such as low resolution, noise interference, and watermarks~\cite{schettini2010underwater}. 
It is essential for detectors to exhibit robustness against these challenges. 
To evaluate the resilience of detectors trained on the WildFake dataset against degraded images, we apply a series of degradation techniques to the images in the testing set:

\begin{itemize}[leftmargin=*]
\item DownSample: Reducing the resolution of high-resolution images to either 128 or 64 pixels.
\item Compression: Introducing compression artifacts by applying JPEG compression with varying quality ratios to the original test images.
\item Geometric Transformation: Randomly flipping or cropping images in the testing set.
\item Watermark: Randomly adding textual or visual watermarks to images at various positions in the testing set.
\item Color Transformation: Randomly altering the brightness, contrast, saturation, and hue of images from the testing set.
\end{itemize}
The results of baseline methods evaluated under these degradation conditions on the WildFake dataset are documented in \cref{tab:robusness_supple}.

Upon analyzing the results, it becomes evident that the ViT-based detector outperforms the ResNet-50-based detector in handling these degraded images. The ResNet-50-based detector is particularly more sensitive to geometric and color transformations, showcasing a higher degree of vulnerability to these specific types of degradation. In contrast, the ViT-based detector exhibits enhanced robustness in the face of such challenges. However, it is noteworthy that both detectors exhibit a reduction in performance when processing images with either image-based or text-based watermarks.
Additionally, the experiments reveal that lower resolutions and lower qualities have a significant impact on the accuracy of both types of detectors. This observation underscores the importance of resolution and image quality in the effective functioning of synthetic image detection models.

{\small
\bibliographystyle{ieeenat_fullname}
\bibliography{egbib}

\begin{thebibliography}{84}
\providecommand{\natexlab}[1]{#1}
\providecommand{\url}[1]{\texttt{#1}}
\expandafter\ifx\csname urlstyle\endcsname\relax
  \providecommand{\doi}[1]{doi: #1}\else
  \providecommand{\doi}{doi: \begingroup \urlstyle{rm}\Url}\fi

\bibitem[Aghasanli et~al.(2023)Aghasanli, Kangin, and
  Angelov]{aghasanli2023interpretable}
Agil Aghasanli, Dmitry Kangin, and Plamen Angelov.
\newblock Interpretable-through-prototypes deepfake detection for diffusion
  models.
\newblock In \emph{Proceedings of the IEEE/CVF International Conference on
  Computer Vision}, pages 467--474, 2023.

\bibitem[AI(2022)]{Laionaesthetics}
LAION AI.
\newblock Laion aesthetics v1. technical report version 1.0, 2022.

\bibitem[Alanov et~al.(2023)Alanov, Titov, Nakhodnov, and
  Vetrov]{alanov2023styledomain}
Aibek Alanov, Vadim Titov, Maksim Nakhodnov, and Dmitry Vetrov.
\newblock Styledomain: Efficient and lightweight parameterizations of stylegan
  for one-shot and few-shot domain adaptation.
\newblock In \emph{Proceedings of the IEEE/CVF International Conference on
  Computer Vision}, pages 2184--2194, 2023.

\bibitem[Antoniou et~al.(2017)Antoniou, Storkey, and Edwards]{antoniou2017data}
Antreas Antoniou, Amos Storkey, and Harrison Edwards.
\newblock Data augmentation generative adversarial networks.
\newblock \emph{arXiv preprint arXiv:1711.04340}, 2017.

\bibitem[Arjovsky et~al.(2017)Arjovsky, Chintala, and
  Bottou]{arjovsky2017wasserstein}
Martin Arjovsky, Soumith Chintala, and L{\'e}on Bottou.
\newblock Wasserstein generative adversarial networks.
\newblock In \emph{ICML}, 2017.

\bibitem[Bao et~al.(2017)Bao, Chen, Wen, Li, and Hua]{bao2017cvae}
Jianmin Bao, Dong Chen, Fang Wen, Houqiang Li, and Gang Hua.
\newblock {CVAE-GAN}: fine-grained image generation through asymmetric
  training.
\newblock In \emph{ICCV}, 2017.

\bibitem[Bird and Lotfi(2023)]{bird2023cifake}
Jordan~J Bird and Ahmad Lotfi.
\newblock Cifake: Image classification and explainable identification of
  ai-generated synthetic images.
\newblock \emph{arXiv preprint arXiv:2303.14126}, 2023.

\bibitem[Brock et~al.(2018)Brock, Donahue, and Simonyan]{brock2018large}
Andrew Brock, Jeff Donahue, and Karen Simonyan.
\newblock Large scale gan training for high fidelity natural image synthesis.
\newblock In \emph{International Conference on Learning Representations}, 2018.

\bibitem[Chang et~al.(2022)Chang, Zhang, Jiang, Liu, and
  Freeman]{chang2022maskgit}
Huiwen Chang, Han Zhang, Lu Jiang, Ce Liu, and William~T Freeman.
\newblock Maskgit: Masked generative image transformer.
\newblock In \emph{Proceedings of the IEEE/CVF Conference on Computer Vision
  and Pattern Recognition}, pages 11315--11325, 2022.

\bibitem[Chang et~al.(2023)Chang, Zhang, Barber, Maschinot, Lezama, Jiang,
  Yang, Murphy, Freeman, Rubinstein, et~al.]{chang2023muse}
Huiwen Chang, Han Zhang, Jarred Barber, AJ Maschinot, Jose Lezama, Lu Jiang,
  Ming-Hsuan Yang, Kevin Murphy, William~T Freeman, Michael Rubinstein, et~al.
\newblock Muse: Text-to-image generation via masked generative transformers.
\newblock \emph{arXiv preprint arXiv:2301.00704}, 2023.

\bibitem[Choi et~al.(2018)Choi, Choi, Kim, Ha, Kim, and Choo]{choi2018stargan}
Yunjey Choi, Minje Choi, Munyoung Kim, Jung-Woo Ha, Sunghun Kim, and Jaegul
  Choo.
\newblock Stargan: Unified generative adversarial networks for multi-domain
  image-to-image translation.
\newblock In \emph{Proceedings of the IEEE conference on computer vision and
  pattern recognition}, pages 8789--8797, 2018.

\bibitem[Choi et~al.(2020)Choi, Uh, Yoo, and Ha]{choi2020stargan}
Yunjey Choi, Youngjung Uh, Jaejun Yoo, and Jung-Woo Ha.
\newblock Stargan v2: Diverse image synthesis for multiple domains.
\newblock In \emph{Proceedings of the IEEE/CVF conference on computer vision
  and pattern recognition}, pages 8188--8197, 2020.

\bibitem[Corvi et~al.(2023)Corvi, Cozzolino, Zingarini, Poggi, Nagano, and
  Verdoliva]{corvi2023detection}
Riccardo Corvi, Davide Cozzolino, Giada Zingarini, Giovanni Poggi, Koki Nagano,
  and Luisa Verdoliva.
\newblock On the detection of synthetic images generated by diffusion models.
\newblock In \emph{ICASSP 2023-2023 IEEE International Conference on Acoustics,
  Speech and Signal Processing (ICASSP)}, pages 1--5. IEEE, 2023.

\bibitem[Deng et~al.(2009)Deng, Dong, Socher, Li, Li, and
  Fei-Fei]{deng2009imagenet}
Jia Deng, Wei Dong, Richard Socher, Li-Jia Li, Kai Li, and Li Fei-Fei.
\newblock Imagenet: A large-scale hierarchical image database.
\newblock In \emph{2009 IEEE conference on computer vision and pattern
  recognition}, pages 248--255. IEEE, 2009.

\bibitem[Dhariwal and Nichol(2021)]{dhariwal2021diffusion}
Prafulla Dhariwal and Alexander Nichol.
\newblock Diffusion models beat gans on image synthesis.
\newblock \emph{Advances in neural information processing systems},
  34:\penalty0 8780--8794, 2021.

\bibitem[Dinh et~al.(2016)Dinh, Sohl-Dickstein, and Bengio]{dinh2016density}
Laurent Dinh, Jascha Sohl-Dickstein, and Samy Bengio.
\newblock Density estimation using real nvp.
\newblock \emph{arXiv preprint arXiv:1605.08803}, 2016.

\bibitem[Doersch(2016)]{doersch2016tutorial}
Carl Doersch.
\newblock Tutorial on variational autoencoders.
\newblock \emph{arXiv preprint arXiv:1606.05908}, 2016.

\bibitem[Dosovitskiy et~al.(2020)Dosovitskiy, Beyer, Kolesnikov, Weissenborn,
  Zhai, Unterthiner, Dehghani, Minderer, Heigold, Gelly,
  et~al.]{dosoViTskiy2020image}
Alexey Dosovitskiy, Lucas Beyer, Alexander Kolesnikov, Dirk Weissenborn,
  Xiaohua Zhai, Thomas Unterthiner, Mostafa Dehghani, Matthias Minderer, Georg
  Heigold, Sylvain Gelly, et~al.
\newblock An image is worth 16x16 words: Transformers for image recognition at
  scale.
\newblock In \emph{International Conference on Learning Representations}, 2020.

\bibitem[Esser et~al.(2021)Esser, Rombach, and Ommer]{esser2021taming}
Patrick Esser, Robin Rombach, and Bjorn Ommer.
\newblock Taming transformers for high-resolution image synthesis.
\newblock In \emph{Proceedings of the IEEE/CVF conference on computer vision
  and pattern recognition}, pages 12873--12883, 2021.

\bibitem[Frank et~al.(2020)Frank, Eisenhofer, Sch{\"o}nherr, Fischer, Kolossa,
  and Holz]{frank2020leveraging}
Joel Frank, Thorsten Eisenhofer, Lea Sch{\"o}nherr, Asja Fischer, Dorothea
  Kolossa, and Thorsten Holz.
\newblock Leveraging frequency analysis for deep fake image recognition.
\newblock In \emph{International conference on machine learning}, pages
  3247--3258. PMLR, 2020.

\bibitem[Gal et~al.(2022)Gal, Alaluf, Atzmon, Patashnik, Bermano, Chechik, and
  Cohen-or]{gal2022image}
Rinon Gal, Yuval Alaluf, Yuval Atzmon, Or Patashnik, Amit~Haim Bermano, Gal
  Chechik, and Daniel Cohen-or.
\newblock An image is worth one word: Personalizing text-to-image generation
  using textual inversion.
\newblock In \emph{The Eleventh International Conference on Learning
  Representations}, 2022.

\bibitem[Goodfellow et~al.(2014)Goodfellow, Pouget-Abadie, Mirza, Xu,
  Warde-Farley, Ozair, Courville, and Bengio]{goodfellow2014generative}
Ian Goodfellow, Jean Pouget-Abadie, Mehdi Mirza, Bing Xu, David Warde-Farley,
  Sherjil Ozair, Aaron Courville, and Yoshua Bengio.
\newblock Generative adversarial nets.
\newblock In \emph{NeurIPS}, 2014.

\bibitem[Gu et~al.(2022{\natexlab{a}})Gu, Meng, Lu, Hou, Minzhe, Liang, Yao,
  Huang, Zhang, Jiang, et~al.]{gu2022wukong}
Jiaxi Gu, Xiaojun Meng, Guansong Lu, Lu Hou, Niu Minzhe, Xiaodan Liang, Lewei
  Yao, Runhui Huang, Wei Zhang, Xin Jiang, et~al.
\newblock Wukong: A 100 million large-scale chinese cross-modal pre-training
  benchmark.
\newblock \emph{Advances in Neural Information Processing Systems},
  35:\penalty0 26418--26431, 2022{\natexlab{a}}.

\bibitem[Gu et~al.(2022{\natexlab{b}})Gu, Chen, Bao, Wen, Zhang, Chen, Yuan,
  and Guo]{gu2022vector}
Shuyang Gu, Dong Chen, Jianmin Bao, Fang Wen, Bo Zhang, Dongdong Chen, Lu Yuan,
  and Baining Guo.
\newblock Vector quantized diffusion model for text-to-image synthesis.
\newblock In \emph{Proceedings of the IEEE/CVF Conference on Computer Vision
  and Pattern Recognition}, pages 10696--10706, 2022{\natexlab{b}}.

\bibitem[Guo et~al.(2023)Guo, Liu, Ren, Grosz, Masi, and
  Liu]{guo2023hierarchical}
Xiao Guo, Xiaohong Liu, Zhiyuan Ren, Steven Grosz, Iacopo Masi, and Xiaoming
  Liu.
\newblock Hierarchical fine-grained image forgery detection and localization.
\newblock In \emph{Proceedings of the IEEE/CVF Conference on Computer Vision
  and Pattern Recognition}, pages 3155--3165, 2023.

\bibitem[He et~al.(2016)He, Zhang, Ren, and Sun]{he2016deep}
Kaiming He, Xiangyu Zhang, Shaoqing Ren, and Jian Sun.
\newblock Deep residual learning for image recognition.
\newblock In \emph{Proceedings of the IEEE conference on computer vision and
  pattern recognition}, pages 770--778, 2016.

\bibitem[He et~al.(2022)He, Chen, Xie, Li, Doll{\'a}r, and
  Girshick]{he2022masked}
Kaiming He, Xinlei Chen, Saining Xie, Yanghao Li, Piotr Doll{\'a}r, and Ross
  Girshick.
\newblock Masked autoencoders are scalable vision learners.
\newblock In \emph{Proceedings of the IEEE/CVF conference on computer vision
  and pattern recognition}, pages 16000--16009, 2022.

\bibitem[hello@civitai.com(2022)]{civitai}
hello@civitai.com.
\newblock civitai, 2022.

\bibitem[Higgins et~al.(2016)Higgins, Matthey, Pal, Burgess, Glorot, Botvinick,
  Mohamed, and Lerchner]{higgins2016beta}
Irina Higgins, Loic Matthey, Arka Pal, Christopher Burgess, Xavier Glorot,
  Matthew Botvinick, Shakir Mohamed, and Alexander Lerchner.
\newblock beta-vae: Learning basic visual concepts with a constrained
  variational framework.
\newblock In \emph{International conference on learning representations}, 2016.

\bibitem[Ho et~al.(2020)Ho, Jain, and Abbeel]{ho2020denoising}
Jonathan Ho, Ajay Jain, and Pieter Abbeel.
\newblock Denoising diffusion probabilistic models.
\newblock \emph{Advances in neural information processing systems},
  33:\penalty0 6840--6851, 2020.

\bibitem[Holub(2017)]{website2023}
Oleksii Holub.
\newblock Discordchatexporter, 2017.

\bibitem[Holub(2022)]{Midjourney}
Oleksii Holub.
\newblock Midjourney, 2022.

\bibitem[Hu et~al.(2021)Hu, Shen, Wallis, Allen-Zhu, Li, Wang, Wang, and
  Chen]{hu2021lora}
Edward~J Hu, Yelong Shen, Phillip Wallis, Zeyuan Allen-Zhu, Yuanzhi Li, Shean
  Wang, Lu Wang, and Weizhu Chen.
\newblock Lora: Low-rank adaptation of large language models.
\newblock \emph{arXiv preprint arXiv:2106.09685}, 2021.

\bibitem[Huang et~al.(2023{\natexlab{a}})Huang, Mao, Chen, and
  Zhang]{huang2023towards}
Mengqi Huang, Zhendong Mao, Zhuowei Chen, and Yongdong Zhang.
\newblock Towards accurate image coding: Improved autoregressive image
  generation with dynamic vector quantization.
\newblock In \emph{Proceedings of the IEEE/CVF Conference on Computer Vision
  and Pattern Recognition}, pages 22596--22605, 2023{\natexlab{a}}.

\bibitem[Huang et~al.(2023{\natexlab{b}})Huang, Mao, Wang, and
  Zhang]{huang2023not}
Mengqi Huang, Zhendong Mao, Quan Wang, and Yongdong Zhang.
\newblock Not all image regions matter: Masked vector quantization for
  autoregressive image generation.
\newblock In \emph{Proceedings of the IEEE/CVF Conference on Computer Vision
  and Pattern Recognition}, pages 2002--2011, 2023{\natexlab{b}}.

\bibitem[Kang et~al.(2023)Kang, Zhu, Zhang, Park, Shechtman, Paris, and
  Park]{kang2023scaling}
Minguk Kang, Jun-Yan Zhu, Richard Zhang, Jaesik Park, Eli Shechtman, Sylvain
  Paris, and Taesung Park.
\newblock Scaling up gans for text-to-image synthesis.
\newblock In \emph{Proceedings of the IEEE/CVF Conference on Computer Vision
  and Pattern Recognition}, pages 10124--10134, 2023.

\bibitem[Karras et~al.(2017)Karras, Aila, Laine, and
  Lehtinen]{karras2017progressive}
Tero Karras, Timo Aila, Samuli Laine, and Jaakko Lehtinen.
\newblock Progressive growing of gans for improved quality, stability, and
  variation.
\newblock \emph{arXiv preprint arXiv:1710.10196}, 2017.

\bibitem[Karras et~al.(2019)Karras, Laine, and Aila]{karras2019style}
Tero Karras, Samuli Laine, and Timo Aila.
\newblock A style-based generator architecture for generative adversarial
  networks.
\newblock In \emph{Proceedings of the IEEE/CVF conference on computer vision
  and pattern recognition}, pages 4401--4410, 2019.

\bibitem[Karras et~al.(2020)Karras, Laine, Aittala, Hellsten, Lehtinen, and
  Aila]{karras2020analyzing}
Tero Karras, Samuli Laine, Miika Aittala, Janne Hellsten, Jaakko Lehtinen, and
  Timo Aila.
\newblock Analyzing and improving the image quality of stylegan.
\newblock In \emph{Proceedings of the IEEE/CVF conference on computer vision
  and pattern recognition}, pages 8110--8119, 2020.

\bibitem[Karras et~al.(2021)Karras, Aittala, Laine, H{\"a}rk{\"o}nen, Hellsten,
  Lehtinen, and Aila]{karras2021alias}
Tero Karras, Miika Aittala, Samuli Laine, Erik H{\"a}rk{\"o}nen, Janne
  Hellsten, Jaakko Lehtinen, and Timo Aila.
\newblock Alias-free generative adversarial networks.
\newblock \emph{Advances in Neural Information Processing Systems},
  34:\penalty0 852--863, 2021.

\bibitem[Kingma and Ba(2014)]{kingma2014adam}
Diederik~P Kingma and Jimmy Ba.
\newblock Adam: A method for stochastic optimization.
\newblock \emph{arXiv preprint arXiv:1412.6980}, 2014.

\bibitem[Kingma and Dhariwal(2018)]{kingma2018glow}
Durk~P Kingma and Prafulla Dhariwal.
\newblock Glow: Generative flow with invertible 1x1 convolutions.
\newblock \emph{Advances in neural information processing systems}, 31, 2018.

\bibitem[Kingma and Welling(2013)]{kingma2013auto}
Diederik~P Kingma and Max Welling.
\newblock Auto-encoding variational bayes.
\newblock \emph{arXiv preprint arXiv:1312.6114}, 2013.

\bibitem[Krizhevsky et~al.(2009)Krizhevsky, Hinton,
  et~al.]{krizhevsky2009learning}
Alex Krizhevsky, Geoffrey Hinton, et~al.
\newblock Learning multiple layers of features from tiny images.
\newblock 2009.

\bibitem[Li et~al.(2023)Li, Chang, Mishra, Zhang, Katabi, and
  Krishnan]{li2023mage}
Tianhong Li, Huiwen Chang, Shlok Mishra, Han Zhang, Dina Katabi, and Dilip
  Krishnan.
\newblock Mage: Masked generative encoder to unify representation learning and
  image synthesis.
\newblock In \emph{Proceedings of the IEEE/CVF Conference on Computer Vision
  and Pattern Recognition}, pages 2142--2152, 2023.

\bibitem[Lin et~al.(2014)Lin, Maire, Belongie, Hays, Perona, Ramanan,
  Doll{\'a}r, and Zitnick]{lin2014microsoft}
Tsung-Yi Lin, Michael Maire, Serge Belongie, James Hays, Pietro Perona, Deva
  Ramanan, Piotr Doll{\'a}r, and C~Lawrence Zitnick.
\newblock Microsoft coco: Common objects in context.
\newblock In \emph{Computer Vision--ECCV 2014: 13th European Conference,
  Zurich, Switzerland, September 6-12, 2014, Proceedings, Part V 13}, pages
  740--755. Springer, 2014.

\bibitem[Lorenz et~al.(2023)Lorenz, Durall, and Keuper]{lorenz2023detecting}
Peter Lorenz, Ricard~L Durall, and Janis Keuper.
\newblock Detecting images generated by deep diffusion models using their local
  intrinsic dimensionality.
\newblock In \emph{Proceedings of the IEEE/CVF International Conference on
  Computer Vision}, pages 448--459, 2023.

\bibitem[Marra et~al.(2018)Marra, Gragnaniello, Cozzolino, and
  Verdoliva]{marra2018detection}
Francesco Marra, Diego Gragnaniello, Davide Cozzolino, and Luisa Verdoliva.
\newblock Detection of gan-generated fake images over social networks.
\newblock In \emph{2018 IEEE conference on multimedia information processing
  and retrieval (MIPR)}, pages 384--389. IEEE, 2018.

\bibitem[Marra et~al.(2019)Marra, Gragnaniello, Verdoliva, and
  Poggi]{marra2019gans}
Francesco Marra, Diego Gragnaniello, Luisa Verdoliva, and Giovanni Poggi.
\newblock Do gans leave artificial fingerprints?
\newblock In \emph{2019 IEEE conference on multimedia information processing
  and retrieval (MIPR)}, pages 506--511. IEEE, 2019.

\bibitem[Nichol et~al.(2022)Nichol, Dhariwal, Ramesh, Shyam, Mishkin, Mcgrew,
  Sutskever, and Chen]{nichol2022glide}
Alexander~Quinn Nichol, Prafulla Dhariwal, Aditya Ramesh, Pranav Shyam, Pamela
  Mishkin, Bob Mcgrew, Ilya Sutskever, and Mark Chen.
\newblock Glide: Towards photorealistic image generation and editing with
  text-guided diffusion models.
\newblock In \emph{International Conference on Machine Learning}, pages
  16784--16804. PMLR, 2022.

\bibitem[OpenAI(2023)]{dalle3}
OpenAI.
\newblock Dall·e 3 system card, 2023.

\bibitem[Parmar et~al.(2018)Parmar, Vaswani, Uszkoreit, Kaiser, Shazeer, Ku,
  and Tran]{parmar2018image}
Niki Parmar, Ashish Vaswani, Jakob Uszkoreit, Lukasz Kaiser, Noam Shazeer,
  Alexander Ku, and Dustin Tran.
\newblock Image transformer.
\newblock In \emph{ICML}, 2018.

\bibitem[Pehlivan et~al.(2023)Pehlivan, Dalva, and
  Dundar]{pehlivan2023styleres}
Hamza Pehlivan, Yusuf Dalva, and Aysegul Dundar.
\newblock Styleres: Transforming the residuals for real image editing with
  stylegan.
\newblock In \emph{Proceedings of the IEEE/CVF Conference on Computer Vision
  and Pattern Recognition}, pages 1828--1837, 2023.

\bibitem[Radford et~al.(2021)Radford, Kim, Hallacy, Ramesh, Goh, Agarwal,
  Sastry, Askell, Mishkin, Clark, et~al.]{radford2021learning}
Alec Radford, Jong~Wook Kim, Chris Hallacy, Aditya Ramesh, Gabriel Goh,
  Sandhini Agarwal, Girish Sastry, Amanda Askell, Pamela Mishkin, Jack Clark,
  et~al.
\newblock Learning transferable visual models from natural language
  supervision.
\newblock In \emph{International conference on machine learning}, pages
  8748--8763. PMLR, 2021.

\bibitem[Rahman et~al.(2023)Rahman, Paul, Sarker, Hakim, and
  Fattah]{rahman2023artifact}
Md~Awsafur Rahman, Bishmoy Paul, Najibul~Haque Sarker, Zaber Ibn~Abdul Hakim,
  and Shaikh~Anowarul Fattah.
\newblock Artifact: A large-scale dataset with artificial and factual images
  for generalizable and robust synthetic image detection.
\newblock \emph{arXiv preprint arXiv:2302.11970}, 2023.

\bibitem[Ramesh et~al.()Ramesh, Dhariwal, Nichol, Chu, and
  Chen]{ramesh2022hierarchical}
Aditya Ramesh, Prafulla Dhariwal, Alex Nichol, Casey Chu, and Mark Chen.
\newblock Hierarchical text-conditional image generation with clip latents.

\bibitem[Razavi et~al.(2019)Razavi, Van~den Oord, and
  Vinyals]{razavi2019generating}
Ali Razavi, Aaron Van~den Oord, and Oriol Vinyals.
\newblock Generating diverse high-fidelity images with vq-vae-2.
\newblock \emph{Advances in neural information processing systems}, 32, 2019.

\bibitem[Rezende and Mohamed(2015)]{rezende2015variational}
Danilo Rezende and Shakir Mohamed.
\newblock Variational inference with normalizing flows.
\newblock In \emph{International conference on machine learning}, pages
  1530--1538. PMLR, 2015.

\bibitem[Rombach et~al.(2022)Rombach, Blattmann, Lorenz, Esser, and
  Ommer]{rombach2022high}
Robin Rombach, Andreas Blattmann, Dominik Lorenz, Patrick Esser, and Bj{\"o}rn
  Ommer.
\newblock High-resolution image synthesis with latent diffusion models.
\newblock In \emph{Proceedings of the IEEE/CVF conference on computer vision
  and pattern recognition}, pages 10684--10695, 2022.

\bibitem[Ruiz et~al.(2023)Ruiz, Li, Jampani, Pritch, Rubinstein, and
  Aberman]{ruiz2023dreambooth}
Nataniel Ruiz, Yuanzhen Li, Varun Jampani, Yael Pritch, Michael Rubinstein, and
  Kfir Aberman.
\newblock Dreambooth: Fine tuning text-to-image diffusion models for
  subject-driven generation.
\newblock In \emph{Proceedings of the IEEE/CVF Conference on Computer Vision
  and Pattern Recognition}, pages 22500--22510, 2023.

\bibitem[Russakovsky et~al.(2015)Russakovsky, Deng, Su, Krause, Satheesh, Ma,
  Huang, Karpathy, Khosla, Bernstein, et~al.]{russakovsky2015imagenet}
Olga Russakovsky, Jia Deng, Hao Su, Jonathan Krause, Sanjeev Satheesh, Sean Ma,
  Zhiheng Huang, Andrej Karpathy, Aditya Khosla, Michael Bernstein, et~al.
\newblock Imagenet: large scale visual recognition challenge.
\newblock \emph{IJCV}, 115\penalty0 (3):\penalty0 211--252, 2015.

\bibitem[Saharia et~al.(2022)Saharia, Chan, Saxena, Li, Whang, Denton,
  Ghasemipour, Gontijo~Lopes, Karagol~Ayan, Salimans,
  et~al.]{saharia2022photorealistic}
Chitwan Saharia, William Chan, Saurabh Saxena, Lala Li, Jay Whang, Emily~L
  Denton, Kamyar Ghasemipour, Raphael Gontijo~Lopes, Burcu Karagol~Ayan, Tim
  Salimans, et~al.
\newblock Photorealistic text-to-image diffusion models with deep language
  understanding.
\newblock \emph{Advances in Neural Information Processing Systems},
  35:\penalty0 36479--36494, 2022.

\bibitem[Sauer et~al.(2022)Sauer, Schwarz, and Geiger]{sauer2022stylegan}
Axel Sauer, Katja Schwarz, and Andreas Geiger.
\newblock Stylegan-xl: Scaling stylegan to large diverse datasets.
\newblock In \emph{ACM SIGGRAPH 2022 conference proceedings}, pages 1--10,
  2022.

\bibitem[Schettini and Corchs(2010)]{schettini2010underwater}
Raimondo Schettini and Silvia Corchs.
\newblock Underwater image processing: state of the art of restoration and
  image enhancement methods.
\newblock \emph{EURASIP journal on advances in signal processing},
  2010:\penalty0 1--14, 2010.

\bibitem[Schuhmann et~al.(2022)Schuhmann, Beaumont, Vencu, Gordon, Wightman,
  Cherti, Coombes, Katta, Mullis, Wortsman, et~al.]{schuhmann2022laion}
Christoph Schuhmann, Romain Beaumont, Richard Vencu, Cade Gordon, Ross
  Wightman, Mehdi Cherti, Theo Coombes, Aarush Katta, Clayton Mullis, Mitchell
  Wortsman, et~al.
\newblock Laion-5b: An open large-scale dataset for training next generation
  image-text models.
\newblock \emph{Advances in Neural Information Processing Systems},
  35:\penalty0 25278--25294, 2022.

\bibitem[Sha et~al.(2022)Sha, Li, Yu, and Zhang]{sha2022fake}
Zeyang Sha, Zheng Li, Ning Yu, and Yang Zhang.
\newblock De-fake: Detection and attribution of fake images generated by
  text-to-image diffusion models.
\newblock \emph{arXiv preprint arXiv:2210.06998}, 2022.

\bibitem[Song et~al.(2020)Song, Meng, and Ermon]{song2020denoising}
Jiaming Song, Chenlin Meng, and Stefano Ermon.
\newblock Denoising diffusion implicit models.
\newblock In \emph{International Conference on Learning Representations}, 2020.

\bibitem[Tao et~al.(2022)Tao, Tang, Wu, Jing, Bao, and Xu]{tao2022df}
Ming Tao, Hao Tang, Fei Wu, Xiao-Yuan Jing, Bing-Kun Bao, and Changsheng Xu.
\newblock Df-gan: A simple and effective baseline for text-to-image synthesis.
\newblock In \emph{Proceedings of the IEEE/CVF Conference on Computer Vision
  and Pattern Recognition}, pages 16515--16525, 2022.

\bibitem[Tao et~al.(2023)Tao, Bao, Tang, and Xu]{tao2023galip}
Ming Tao, Bing-Kun Bao, Hao Tang, and Changsheng Xu.
\newblock Galip: Generative adversarial clips for text-to-image synthesis.
\newblock In \emph{Proceedings of the IEEE/CVF Conference on Computer Vision
  and Pattern Recognition}, pages 14214--14223, 2023.

\bibitem[Van Den~Oord et~al.(2017)Van Den~Oord, Vinyals, et~al.]{van2017neural}
Aaron Van Den~Oord, Oriol Vinyals, et~al.
\newblock Neural discrete representation learning.
\newblock \emph{Advances in neural information processing systems}, 30, 2017.

\bibitem[Verdoliva et~al.()Verdoliva, Cozzolino, and Nagano]{verdoliva2022}
Luisa Verdoliva, Davide Cozzolino, and Koki Nagano.
\newblock 2022 ieee image and video processing cup synthetic image detection.

\bibitem[Voynov et~al.(2023)Voynov, Chu, Cohen-Or, and Aberman]{voynov2023p+}
Andrey Voynov, Qinghao Chu, Daniel Cohen-Or, and Kfir Aberman.
\newblock $ p+ $: Extended textual conditioning in text-to-image generation.
\newblock \emph{arXiv preprint arXiv:2303.09522}, 2023.

\bibitem[Wang et~al.(2019)Wang, Wang, Owens, Zhang, and
  Efros]{wang2019detecting}
Sheng-Yu Wang, Oliver Wang, Andrew Owens, Richard Zhang, and Alexei~A Efros.
\newblock Detecting photoshopped faces by scripting photoshop.
\newblock In \emph{Proceedings of the IEEE/CVF International Conference on
  Computer Vision}, pages 10072--10081, 2019.

\bibitem[Wang et~al.(2020)Wang, Wang, Zhang, Owens, and Efros]{wang2020cnn}
Sheng-Yu Wang, Oliver Wang, Richard Zhang, Andrew Owens, and Alexei~A Efros.
\newblock Cnn-generated images are surprisingly easy to spot... for now.
\newblock In \emph{Proceedings of the IEEE/CVF conference on computer vision
  and pattern recognition}, pages 8695--8704, 2020.

\bibitem[Wang et~al.(2023)Wang, Bao, Zhou, Wang, Hu, Chen, and
  Li]{wang2023dire}
Zhendong Wang, Jianmin Bao, Wengang Zhou, Weilun Wang, Hezhen Hu, Hong Chen,
  and Houqiang Li.
\newblock Dire for diffusion-generated image detection.
\newblock \emph{arXiv preprint arXiv:2303.09295}, 2023.

\bibitem[Wang et~al.(2022)Wang, Montoya, Munechika, Yang, Hoover, and
  Chau]{wang2022diffusiondb}
Zijie~J Wang, Evan Montoya, David Munechika, Haoyang Yang, Benjamin Hoover, and
  Duen~Horng Chau.
\newblock Diffusiondb: A large-scale prompt gallery dataset for text-to-image
  generative models.
\newblock \emph{arXiv preprint arXiv:2210.14896}, 2022.

\bibitem[Wu et~al.(2023)Wu, Zhou, and Zhang]{wu2023generalizable}
Haiwei Wu, Jiantao Zhou, and Shile Zhang.
\newblock Generalizable synthetic image detection via language-guided
  contrastive learning.
\newblock \emph{arXiv preprint arXiv:2305.13800}, 2023.

\bibitem[Yeh et~al.(2023)Yeh, Hsieh, Gao, Yang, Oh, and
  Gong]{yeh2023navigating}
Shin-Ying Yeh, Yu-Guan Hsieh, Zhidong Gao, Bernard~BW Yang, Giyeong Oh, and
  Yanmin Gong.
\newblock Navigating text-to-image customization: From lycoris fine-tuning to
  model evaluation.
\newblock \emph{arXiv preprint arXiv:2309.14859}, 2023.

\bibitem[Young et~al.(2014)Young, Lai, Hodosh, and Hockenmaier]{young2014image}
Peter Young, Alice Lai, Micah Hodosh, and Julia Hockenmaier.
\newblock From image descriptions to visual denotations: New similarity metrics
  for semantic inference over event descriptions.
\newblock \emph{Transactions of the Association for Computational Linguistics},
  2:\penalty0 67--78, 2014.

\bibitem[Yu et~al.(2015)Yu, Seff, Zhang, Song, Funkhouser, and
  Xiao]{yu2015lsun}
Fisher Yu, Ari Seff, Yinda Zhang, Shuran Song, Thomas Funkhouser, and Jianxiong
  Xiao.
\newblock Lsun: Construction of a large-scale image dataset using deep learning
  with humans in the loop.
\newblock \emph{arXiv preprint arXiv:1506.03365}, 2015.

\bibitem[Zhang et~al.(2019)Zhang, Huang, Li, Zhao, and Zhang]{ZhangHLZ019}
Jianfu Zhang, Yuanyuan Huang, Yaoyi Li, Weijie Zhao, and Liqing Zhang.
\newblock Multi-attribute transfer via disentangled representation.
\newblock In \emph{AAAI}, 2019.

\bibitem[Zhang et~al.(2023)Zhang, Rao, and Agrawala]{zhang2023adding}
Lvmin Zhang, Anyi Rao, and Maneesh Agrawala.
\newblock Adding conditional control to text-to-image diffusion models.
\newblock In \emph{Proceedings of the IEEE/CVF International Conference on
  Computer Vision}, pages 3836--3847, 2023.

\bibitem[Zhu et~al.(2017)Zhu, Park, Isola, and Efros]{zhu2017unpaired}
Jun-Yan Zhu, Taesung Park, Phillip Isola, and Alexei~A Efros.
\newblock Unpaired image-to-image translation using cycle-consistent
  adversarial networks.
\newblock In \emph{Proceedings of the IEEE international conference on computer
  vision}, pages 2223--2232, 2017.

\bibitem[Zhu et~al.(2023)Zhu, Chen, Yan, Huang, Lin, Li, Tu, Hu, Hu, and
  Wang]{zhu2023genimage}
Mingjian Zhu, Hanting Chen, Qiangyu Yan, Xudong Huang, Guanyu Lin, Wei Li,
  Zhijun Tu, Hailin Hu, Jie Hu, and Yunhe Wang.
\newblock Genimage: A million-scale benchmark for detecting ai-generated image.
\newblock \emph{arXiv preprint arXiv:2306.08571}, 2023.

\end{thebibliography}
}

\end{document}